%% file: po-main.tex
\documentclass[sigconf]{acmart}
\AtBeginDocument{%
  }

\setcopyright{acmlicensed}
\copyrightyear{2018}
\acmYear{2018}
\acmDOI{XXXXXXX.XXXXXXX}
\acmConference[Conference acronym 'XX]{Make sure to enter the correct
  conference title from your rights confirmation email}{June 03--05,
  2018}{Woodstock, NY}
\acmISBN{978-1-4503-XXXX-X/2018/06}

\usepackage{todonotes}
\usepackage{listings}
\usepackage{geometry}
\usepackage{adjustbox}
\usepackage{breqn}
\usepackage{amsmath}
\usepackage{amsfonts}
\usepackage{tcolorbox}
\tcbuselibrary{skins,breakable}
\usepackage{enumitem}
\usepackage{algorithm}
\usepackage[noend]{algpseudocode}
\usepackage{enumitem}  
\usepackage{caption}

\newcommand{\protegi}{\texttt{ProTeGi}~}
\newcommand{\ape}{\texttt{APE}~}
\newcommand{\apeopro}{\texttt{APE-OPRO}~}
\newcommand{\opro}{\texttt{OPRO}~}

\begin{document}

\title{Prompt Smart, Pay Less: Cost-Aware APO for Real-World Applications}

\author{Jayesh Choudhari}
\authornote{Both authors contributed equally to this research.}
\affiliation{%
  \institution{Viator, TripAdvisor}
  \city{London}
  \country{UK}
}

\author{Piyush Kumar Singh}
\authornotemark[1]
\affiliation{%
  \institution{Viator, TripAdvisor}
  \city{London}
  \country{UK}
}

\author{Douglas McIlwraith}
\affiliation{%
  \institution{Viator, TripAdvisor}
  \city{London}
  \country{UK}
}

\author{Snehal Nair}
\affiliation{%
  \institution{Viator, TripAdvisor}
  \city{London}
  \country{UK}
}


\renewcommand{\shortauthors}{Choudhari, Singh et al.}


\input{abstract}


\received{20 February 2007}
\received[revised]{12 March 2009}
\received[accepted]{5 June 2009}

\maketitle
\input{intro}
\input{problem_formulation}

\input{cost}
\input{expt-setup}
\input{results-discussions}
\input{related_work}
\input{conclusion}


\begin{acks}
We gratefully acknowledge the valuable suggestions and consistent feedback provided by \href{https://www.linkedin.com/in/karthikn5/}{Karthik Nagesh}.
\end{acks}

\bibliographystyle{ACM-Reference-Format}
\bibliography{po-bib}

\appendix
\input{appendix-all}

\end{document}

%% file: abstract.tex
\begin{abstract}
Prompt design is a critical factor in the effectiveness of Large Language Models (LLMs), yet remains largely heuristic, manual, and difficult to scale. 
This paper presents the first comprehensive evaluation of Automatic Prompt Optimization (APO) methods for real-world, high-stakes multiclass classification in a commercial setting, addressing a critical gap in the existing literature where most of the APO frameworks have been validated only on benchmark classification tasks of limited complexity. 


We introduce APE-OPRO, a novel hybrid framework that combines the complementary strengths of APE and OPRO, 
achieving notably better cost-efficiency, around $18\%$ improvement over OPRO, without sacrificing performance.
We benchmark APE-OPRO alongside both gradient-free (APE, OPRO) and gradient-based (ProTeGi) methods on a dataset of ~2,500 labeled products.

Our results highlight key trade-offs: ProTeGi offers the strongest absolute performance at lower API cost but higher computational time as noted in~\cite{protegi}, while APE-OPRO strikes a compelling balance between performance, API efficiency, and scalability. 
We further conduct ablation studies on depth and breadth hyperparameters, and reveal notable sensitivity to label formatting, indicating implicit sensitivity in LLM behavior.
These findings provide actionable insights for implementing APO in commercial applications and establish a foundation for future research in multi-label, vision, and multimodal prompt optimization scenarios.
\end{abstract}

%% file: intro.tex
\begin{figure}[H]
    \centering
    \includegraphics[width=0.95\linewidth]{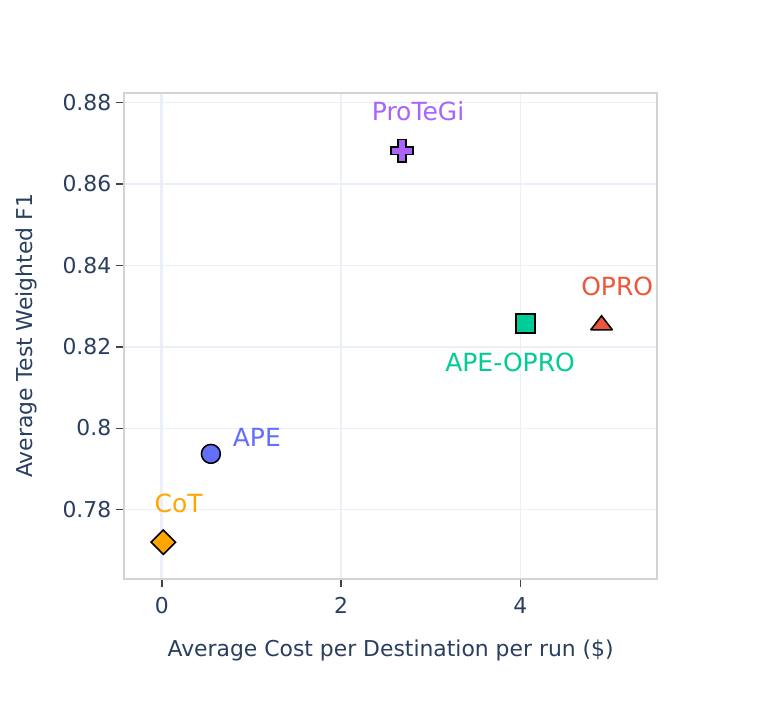}
    \caption{Comparison of evaluated methods based on Mean Test Weighted F1 and Mean Cost per destination. Results are averaged over five runs per destination, spanning 2,500 products across
10 destinations and each run involves an optimization process with 10 iterations and 10 prompts per iteration using GPT-4.1 (optimizer) and GPT-4o-mini (scorer). See Section~\ref{sec:cost_computation} for details on cost computation.}
    \label{fig:costvsperformance}
\end{figure}
\section{Introduction}
Large Language Models (LLMs) have revolutionized natural language processing across tasks such as summarization, question answering, and classification. 
However, when used in a prompt-based setting, their real-world performance is highly sensitive to the phrasing and structure of input prompts.
Designing effective prompts is often a manual, heuristic-driven process that lacks reproducibility and scalability. Zero-shot prompting, while attractive for its simplicity, frequently falls short of production-grade performance without prompt optimization.
To address the challenge of effective prompt design, recent studies have introduced automatic prompt optimization (APO) techniques that iteratively refine prompts using model feedback or search strategies. Although these approaches have shown promising results, they have primarily been evaluated on tasks such as binary classification or short-form output, often using benchmark datasets with limited complexity. 
To the best of our knowledge, this work represents the first comprehensive evaluation of APO methods in a complex, real-world single-label multiclass classification setting. 


Building on this foundation, our study presents a systematic investigation of prompt optimization strategies for a commercially impactful classification task at Viator (a global platform that aggregates and sells curated travel experiences). 
We evaluate a range of APO methods and propose a novel hybrid approach APE-OPRO, that integrates the strengths of APE's exploration with OPRO's metaprompt-based approach, and have performed a detailed cost-performance analysis on a curated dataset of 2,500 manually labeled products in the top 10 globally diverse destinations. 
The techniques explored in this work fall into two main approaches:

\begin{itemize}[wide, labelwidth=!, labelindent=0pt, nosep]
    \item \textbf{Gradient-Free Methods} have gained significant attention in recent literature, with several promising approaches such as Dual-Phase Accelerated Prompt Optimization~\cite{yang2024dual}, GPO~\cite{GPO}, and PE2~\cite{PE2}. In this work, we focus on two widely adopted gradient-free methods—APE~\cite{APE} and OPRO~\cite{OPRO}. These serve as strong baselines and exemplify contrasting strategies for prompt optimization in the absence of gradient information.

    \item \textbf{Gradient-Based Methods} have also attracted substantial interest, with approaches such as ProTeGi~\cite{protegi}, TextGrad~\cite{textgrad}, and GREATER~\cite{das2024greater} advancing differentiable prompt optimization. In this study, we specifically focus on ProTeGi~\cite{protegi}, which offers a robust and scalable gradient-based framework that aligns well with our experimental design and evaluation objectives.
\end{itemize}
~\\
In addition to gradient-based and gradient-free strategies, there exists a third category of optimization methods inspired by evolutionary algorithms. Evolutionary methods explore the prompt search space using biologically inspired mechanisms such as mutation, selection, and recombination. Notable examples include EvoPrompt~\cite{tong2025evoprompt}, PromptBreeder~\cite{fernando2023promptbreeder} and SPRIG~\cite{zhang2024sprig}. As noted in the OPRO paper~\cite{OPRO}, EvoPrompt’s
optimization trajectory tends to be less stable compared to OPRO,
primarily due to its reliance on limited prompt history and the
absence of task exemplars during refinement. Consequently, we do
not include these methods in our experimental evaluation.

Beyond these methods, there also exist alternative optimization strategies that do not fit neatly into the gradient-based, gradient-free, or evolutionary frameworks. For example, PromptAgent~\cite{Promptagent} employs Monte Carlo Tree Search to explore and select optimal prompt sequences. CRISPO~\cite{he2025crispo} and SCULPT~\cite{kumar2024sculpt} rely on iterative refinement guided by model generated critiques and suggestions. PREFER~\cite{zhang2023prefer} adopts an ensemble based feedback-reflect-refine loop to enhance prompt quality. While promising, these strategies are considered complementary and are not included in our current evaluation.

\begin{figure*}[ht]
    \centering
\includegraphics[width=0.88\linewidth]{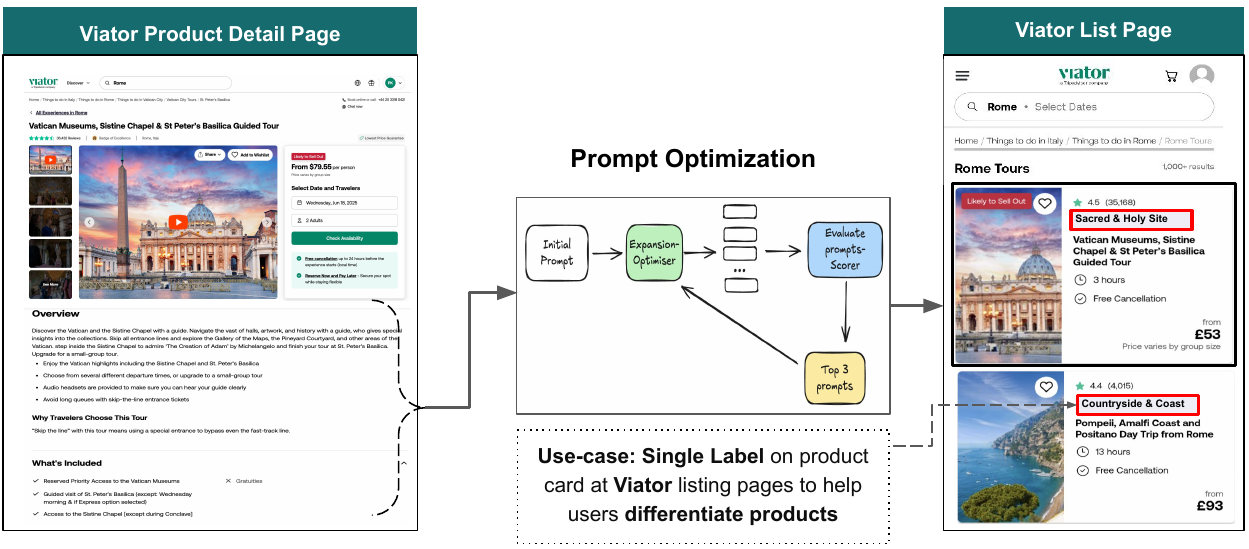}
    \caption{Workflow for single-label classification of Viator products. Product text from Viator’s product description pages is passed to a prompt optimization method, which generates a single, destination-specific label to improve product differentiation on listing pages.}
    \label{fig:viator_labeling_usecase}
\end{figure*}

\subsection{Contributions}

Our work makes the following key contributions:
\begin{itemize}
    \item \textbf{Pushing Beyond Binary: Real-World Evaluation of APO in Multiclass Classification}: We introduce a challenging real-world task of single-label, multiclass classification using our proprietary dataset of 2,500 high-revenue products across ten globally diverse destinations. This departs from the binary classification focus prevalent in most prior work, directly addressing the need for such evaluation highlighted
    as one of the limitations in \cite{protegi}.    
    \item \textbf{APE-OPRO: A hybrid APO framework}: We propose APE-OPRO, a novel hybrid approach that combines APE~\cite{APE} and OPRO~\cite{OPRO}. APE-OPRO matches OPRO's performance while significantly reducing API costs by $\sim\!18\%$ as shown in figure~\ref{fig:costvsperformance}.
    \item \textbf{Cost-Performance Trade-offs in Prompt Optimization}: We perform a comprehensive cost-sensitive analysis of diverse APO approaches including gradient-based (ProTeGi) and gradient-free (APE, OPRO) and highlight their trade-offs between weighted F1 performance and API cost. Notably, ProTeGi delivers strong cost efficiency despite requiring longer execution time compared to other algorithms.
    \item \textbf{Revealing Implicit LLM sensitivity}: We uncover a previously unreported sensitivity in APE where prompt performance is sensitive to label formatting, emphasizing how the stochastic behavior of large language models can impact automated prompt optimization.
\end{itemize}

Although this study focuses on single label classification, we outline a generalizable framework for cost-aware APO evaluation and identify promising research frontiers in multi-label, multimodal, and multilingual prompt optimization.

%% file: problem_formulation.tex
\section{Problem Formulation}
To improve product discoverability on Viator, we initially adopted a multi-label taxonomy in which individual products could be assigned multiple, overlapping categories (e.g., a tour labeled both ‘Historical’ and ‘Cultural’). 
While this approach improved overall visibility, it also caused popular products to appear in many categories, which reduced the diversity of products shown to users and limited exposure for less prominent yet relevant offerings.
To address this issue, we transitioned to a single label, multiclass taxonomy, where each product is assigned one mutually exclusive label from a predefined set. This structure reduces cognitive load and facilitates easier comparison across distinct categories. In contrast to traditional hierarchical systems, our flat, destination-specific taxonomy is better aligned with how users naturally explore travel experiences and contributes to improved product discoverability, as illustrated in Figure~\ref{fig:viator_labeling_usecase}.

We frame this task as a single-label classification problem specific to each destination $d$ , where, for a given product description $x \in \mathcal{D}$, the objective is to assign the most appropriate label from a defined, mutually exclusive set  $\mathcal{C}_d = \{ c_1^d, c_2^d, \ldots, c_{N_d}^d \}$ .

Rather than relying on conventional supervised training approaches due to limited ground truth, we frame the problem as a prompt optimization task using Large Language Models.
Our goal is to find an optimal natural language prompt \( p^* \in \mathcal{P} \) such that the LLM, when conditioned on \( p^* \) and input \( x \), produces the correct label. 
Formally, we seek:
\begin{equation}
p^* = \arg\max_{p \in \mathcal{P}}  \mathbb{E}_{\mathcal{D}' \sim \mathcal{D}} \left[ f(p, \mathcal{D}') \right],    
\end{equation}
where $f(p, \mathcal{D}')$  is an evaluation metric (e.g., weighted f1) calculated over a sample  $\mathcal{D}'$ of the training data, assessing the LLM performance using prompt $p$.

\begin{figure}
    \centering
    \includegraphics[width=0.85\linewidth]{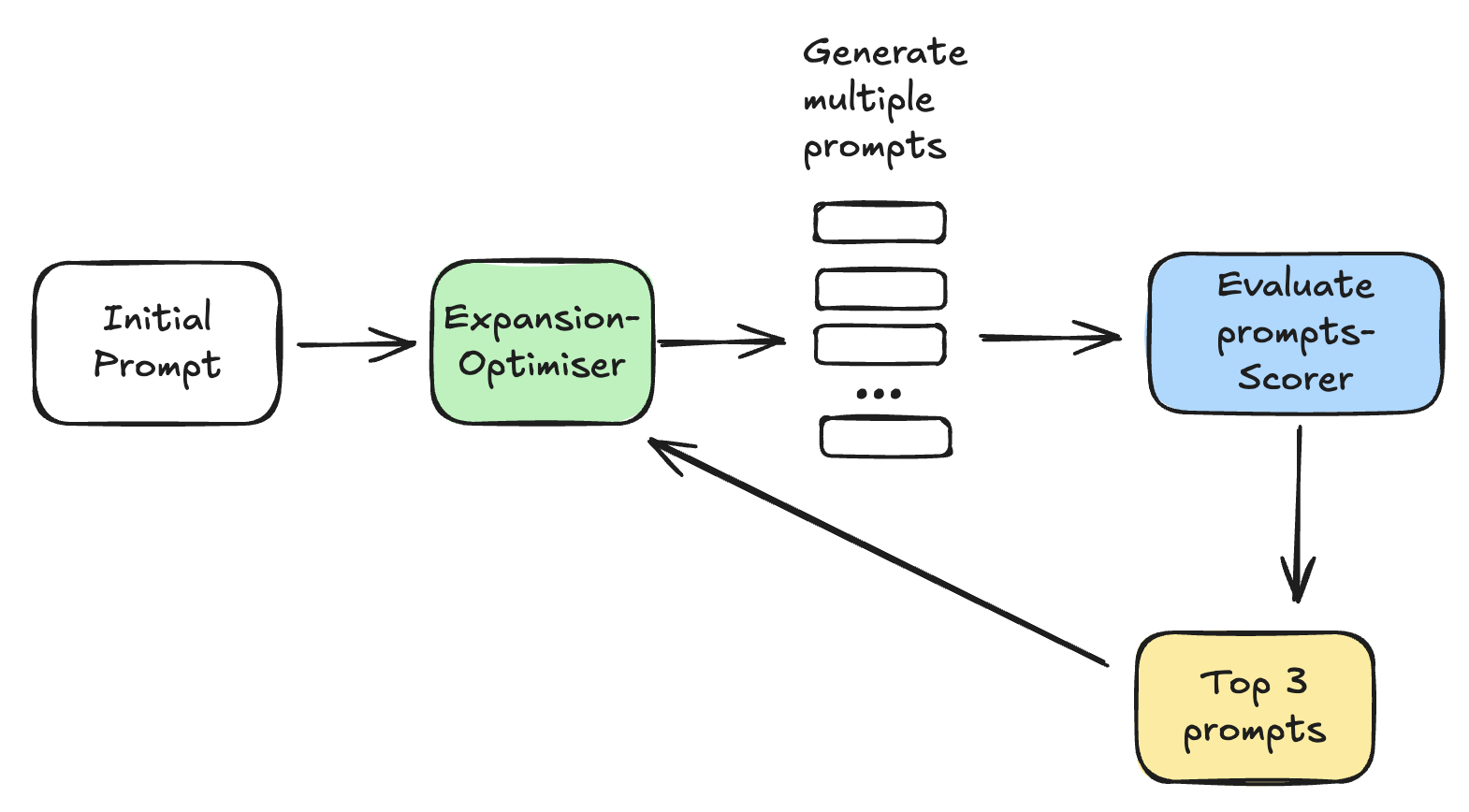}
    \caption{Prompt Optimization Framework}
    \label{fig:prompt_optimisation_framework}
\end{figure}

To solve this optimization problem, we investigate several APO techniques, categorizing them as gradient-free (APE \cite{APE}, OPRO \cite{OPRO}) and gradient-based (ProTeGi \cite{protegi}). We also introduce a novel hybrid method, APE-OPRO, combining APE's initialization with OPRO's metaprompt guided prompt generation. We unify these methods within a general prompt optimization framework as described in Figure~\ref{fig:prompt_optimisation_framework}.

This framework operates iteratively, starting with an initial prompt. Each iteration involves an \textbf{expansion phase}, where an \textit{ optimizer model} generates multiple candidate prompts (defining the \textbf{width}). These candidates are then evaluated using a computationally less expensive \textit{scorer model}. The top ($k = 3$) performing prompts, are selected and fed back into the optimizer for the next iteration. 
This cycle continues for a fixed number of iterations, defined as the \textbf{depth}. The framework is flexible enough to encapsulate the explored APO strategies. 
This modular design, which separates prompt generation from evaluation, facilitates experimentation across different techniques.


Each destination uses a custom template for the prompt where the candidate label set \( \mathcal{C}_d \) is inserted into a fixed template~\ref{sec:base-system-prompt} to preserve semantic relevance and improve performance.

%% file: cost.tex
\section{Cost Computation}
\label{sec:cost_computation}

We compute the cost of APO for each destination as follows:
\begin{itemize}[wide, labelwidth=!, labelindent=0pt, nosep]
    \item $N$ be the total number of iterations,
    \item $M$ be the number of prompts generated per iteration,
    \item $C_{\text{in}}^{\text{op}}(i)$ and $C_{\text{out}}^{\text{op}}(i)$ be the input and output token costs for the optimizer model (expansion phase) in iteration $i$ (prompt generation),
    \item $C_{\text{in}}^{\text{sc}}(i, j)$ and $C_{\text{out}}^{\text{sc}}(i, j)$ be the input and output token costs for scoring (scoring phase) the $j$-th prompt in iteration $i$.
\end{itemize}

Then, the total cost\footnote{Reported costs do not account for potential savings from prompt caching or batch API calls in offline scenarios} for one run is:
\begin{equation}
\label{eq:cost-equation}
\text{Total Cost} = \sum_{i=1}^{N} \left[ C_{\text{in}}^{\text{op}}(i) + C_{\text{out}}^{\text{op}}(i) + \sum_{j=1}^{M} \left( C_{\text{in}}^{\text{sc}}(i, j) + C_{\text{out}}^{\text{sc}}(i, j) \right) \right]
\end{equation}

For each method (APE, OPRO, APE-OPRO, and ProTeGi), the total cost is computed as the cumulative sum of input and output tokens across all optimization iterations, as defined in Equation~\ref{eq:cost-equation}. As shown in Figure~\ref{fig:prompt_optimisation_framework}, each iteration includes two stages: prompt expansion (handled by optimizer model) and prompt evaluation (handled by scorer model). Unless otherwise stated, we use GPT-4.1 for optimization and GPT-4o-mini for scoring\footnote{https://openai.com/api/pricing/}. 

%% file: expt-setup.tex
\section{Experiment Setup}
\subsection{Dataset}
Our dataset comprises proprietary Viator travel experiences from ten diverse global destinations, selected to capture a wide range of user interests and destination types. These include major European cities (Rome, Paris, Lisbon, Amsterdam, Edinburgh, Athens), Asia (Tokyo), North American entertainment centers (Las Vegas), and island or remote destinations (Maui, Reykjavik), offering geographic and cultural diversity.

\begin{table}[ht]
  \caption{Overview of the dataset split by destination, including train/test size, number of unique labels, and average token count per product description}
  \label{tab:dataset}
  \centering
  \begin{tabular}{lcccc}
    \toprule
    \textbf{Destination} & \textbf{Train} & \textbf{Test} & \textbf{\#Labels} & \textbf{Avg. \#Tokens} \\
    \midrule
    Rome         & 52  & 390 & 14 & 190 \\
    Lisbon       & 75  & 213 & 23 & 180 \\
    Amsterdam    & 50  & 109 & 16 & 185 \\
    Paris        & 73  & 229 & 19 & 189 \\
    Las Vegas    & 59  & 186 & 17 & 179 \\
    Athens       & 58  & 185 & 19 & 198 \\
    Tokyo        & 65  & 190 & 18 & 183  \\
    Edinburgh    & 51  & 125 & 15 & 179 \\
    Reykjavik    & 63  & 114 & 17 & 183 \\
    Maui         & 48  & 166 & 12 & 173 \\
    \midrule
  \end{tabular}
\end{table}
\begin{figure}
    \centering
    \includegraphics[width=0.95\linewidth]{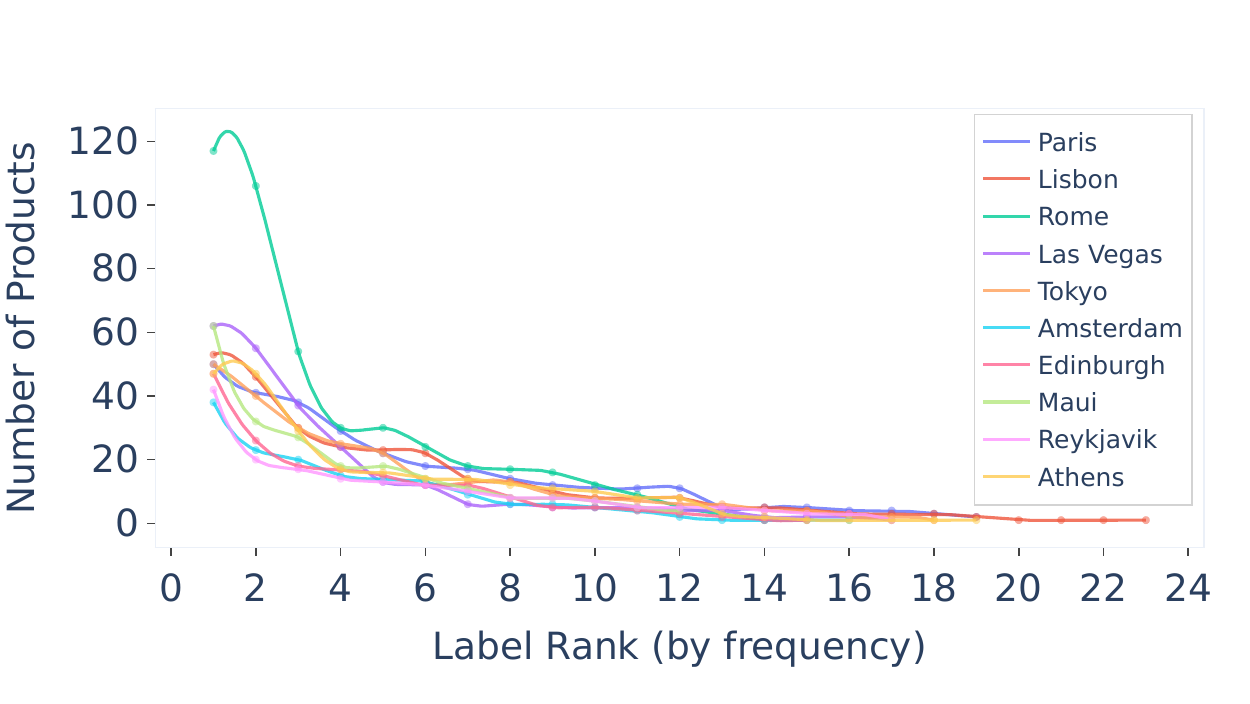}
    \caption{Distribution of labels in Top 10 Destinations}
    \label{fig:label-dist-top10-dest}
\end{figure}

\subsubsection{\textbf{Data Collection and Preprocessing}}
For each destination, we extracted product information from Viator's product description as illustrated in Figure~\ref{fig:viator_labeling_usecase}. Each entry in the experience contains an average of 185 tokens of descriptive text.

\subsubsection{\textbf{Labeling Process}}
To initiate the taxonomy refinement, we manually labeled a subset of top ranked products that cover 80\% of the revenue within each destination, focusing on popular and representative experiences. This revenue-weighted sampling approach ensures that our taxonomy captures economically significant experience categories while maintaining destination-specific relevance. Two domain experts independently labeled each experience, achieving an interannotator agreement of 87. 3\% (Cohen's $\kappa$ = 0.81), with disagreements resolved by discussion.

We developed a flat labeling taxonomy (a non-hierarchical set of 66 mutually exclusive categories) to classify travel experience products across 10 major destinations. The number of labels varies by destination (from 12 for Maui to 23 for Lisbon), reflecting the diversity of available experiences. Some labels are destination-specific (e.g., ``Volcano Tours'' for Reykjavik), while others are universal (e.g., ``City Highlights'').\footnote{Due to proprietary constraints, the dataset is not publicly released but can be simulated using category-rich classification tasks.}

\subsubsection{\textbf{Dataset Characteristics}}
As illustrated in Figure \ref{fig:label-dist-top10-dest}, the distribution of labels follows a long tail pattern, where a few labels represent a large proportion of products, typically representing highly popular and destination-specific activities, while many other labels correspond to niche or very common offerings. For example, in Rome, ``Sacred \& Holy Site'' accounts for 26.47\% and ``Countryside \& Coast'' represents 8.82\% of all experiences, while in Las Vegas,  ``Adventure'' represents 19\%, ``City Highlight'' represents 3.24\%.

The ``Other Experiences'' label is included as an additional candidate label across all destinations during the training stage to handle products associated with rare or previously unseen categories. When the label is assigned, the product is flagged for manual review by Destination Managers, who periodically assess whether a new dedicated label should be introduced based on product volume.

\subsubsection{\textbf{Destination Selection Criteria for Evaluation}}
To provide a representative evaluation of prompt optimization methods, we focus on three key destinations: Rome, Amsterdam, and Lisbon. These were chosen to capture a range of product diversity and complexity in the labeling. Rome represents a destination with well defined tourist attractions and relatively clear categorization.  In contrast, Amsterdam offers moderate complexity due to its diverse range of experience types, while Lisbon presents the most challenging case with the highest number of distinct labels (23).

\subsection{Train-Test Split}
To ensure consistency across methods, we adopt a standardized
train-test split strategy. We perform stratified sampling by randomly sampling up to
four examples per label within each destination to construct the
training set. The remaining examples are reserved for evaluation in the
test set. Using four examples per label ensures sufficient representation during training while also minimizing the risk of labels appearing only in the training set and not in the test set, leading to a fairer evaluation.
All methods are trained exclusively on the training set and the test set is strictly held out and used only for final evaluation. The number of training and test instances after the split is summarized in Table~\ref{tab:dataset}.

\subsection{Prompt Optimization Configuration}
All methods are executed under a uniform configuration of hyperparameters unless otherwise stated. For APE, OPRO, and APE-OPRO, we fix the optimization depth (number of iterations) to 10, and the breadth (number of prompts generated per iteration) to 10.

For ProTeGi, we similarly set the \textit{search-depth} to 10, aligning it with the other methods for comparability. The \textit{max-expansion-factor}, which controls the number of prompt candidates passed to the selection stage in each iteration, is also set to 10. This parameter plays a role analogous to the breadth in the other methods. All remaining hyperparameters for ProTeGi are set as described in Section~3 (Setup) of the original paper, except for the subset $\mathcal{D}_{mini}$, which we set to half of the training data, i.e., $\mathcal{D}_{train/2}$.

Prompt selection at each iteration is based on macro F1 score, which encourages generalizable prompts by evaluating performance uniformly across all classes.
We do not incorporate any early stopping criteria in our experiments. Instead, each algorithm is allowed to run for a fixed maximum number of iterations, ensuring a consistent evaluation framework across all methods.

\subsection{Evaluation Metric}
\label{sec:evaluation_metric}
For final performance evaluation, we report the weighted F1 score due to the high class imbalance present in our multiclass classification task. While macro F1 offers equal weighting across classes, it is highly sensitive to errors in rare labels, where a single misclassification can cause the F1 score for that class to drop to zero. This can disproportionately skew the overall result.

In contrast, weighted F1 accounts for class frequency, providing a more robust and representative measure of model performance in practical scenarios—particularly where dominant classes are of primary interest. A detailed definition of this metric is provided in Appendix~\ref{sec:metrics}.

\subsection{Algorithms}
\label{sec:algorithms}
\subsubsection{\textbf{CoT}\cite{wei2022chain}}
This serves as the baseline Chain-of-Thought (CoT) prompt, in which the phrase - \textit{Think step by step before answering} is appended to the base instruction. We include this standard CoT formulation to establish a performance baseline, enabling a comparison of the cost-performance trade-offs associated with more advanced prompting strategies.

\subsubsection{\textbf{APE}\cite{APE}}
Among various prompt optimization strategies, the APE framework is notable for its conceptual simplicity and minimal explicit feedback. We adapt APE following the framework in figure \ref{fig:prompt_optimisation_framework} for our use case as described in Algorithm \ref{alg:ape-framework}.



\begin{algorithm}
\caption{APE-style Prompt Optimization Framework}
\label{alg:ape-framework}
\begin{algorithmic}[1]
\State \textbf{Input:} Initial prompt $p_0$, training data $\mathcal{D}_{train}$, \#iterations $T$, \#prompts $g = 10$
\State Initialize $S_0 = \{p_0\}$, $i = 0$
\While{$i < T$}
    \State $i \gets i + 1$; $S_i = \emptyset$
    \For{each prompt $p \in S_{i-1}$}
        \State Generate $g / |S_{i-1}|$ semantically similar prompts using optimizer \Comment{Section~\ref{sec:gen-similar-prompt}}
        \State Add generated prompts to $S_i$
    \EndFor
    \State Compute macro F1 for each prompt in $S_i$ (using scorer) on $\mathcal{D}_{train}$
    \State Select top $3$ performing prompts $\rightarrow S_i$
\EndWhile
\State \textbf{Output:} Best prompt (based on macro F1) $p_T$ over all iterations
\end{algorithmic}
\end{algorithm}
\subsubsection{\textbf{OPRO}\cite{OPRO}}
We adapt the original OPRO framework to better suit our task of iterative prompt refinement for label definition in prompt based classification. Specifically, we make two key modifications: (1) we exclude exemplars from the metaprompt, as our generated prompts are significantly longer ($\sim\!1300$) tokens on average) and more descriptive than standard OPRO setups; and (2) we restrict the metaprompt to the top 3 performing prompts with their scores, rather than the original set of 20, to maintain clarity and reduce prompt length. The meta-prompt template is provided in Section~\ref{sec:opro-metaprompt-template}.

\subsubsection{\textbf{APE-OPRO}}
In this method, we integrate the methodologies of APE and OPRO, to improve the prompt optimization for our specific use case. Since OPRO does not incorporate any prompt and scores in its initial iteration, we use APE’s initialization strategy by generating semantically similar variants of the initial system prompt~\ref{sec:base-system-prompt}. This step mirrors the Iterative Monte Carlo Search approach in APE \cite{APE}. We then evaluate these candidate prompts on a training dataset and select the top 3 prompts based on macro F1. These high-performing prompts, along with their evaluation scores, are embedded into a metaprompt~\ref{sec:opro-metaprompt-template}, which is used to guide the subsequent round of prompt generation. This iterative process is repeated for 10 iterations (including APE's initial iteration) to progressively refine and improve prompt quality.

\subsubsection{\textbf{ProTeGi}\cite{protegi}}
\protegi (Prompt Tuning via Gradient-Inspired Optimization) treats prompt engineering as an iterative process guided by performance feedback, mimicking gradient descent through natural language updates. 
As shown in figure~\ref{fig:prompt_optimisation_framework}, we adapt ProTeGi to suit our specific task as described in Algorithm \ref{alg:protegi-framework}.
\begin{algorithm}
\caption{ProTeGi Prompt Optimization Framework}
\label{alg:protegi-framework}
\begin{algorithmic}[1]
\State \textbf{Input:} Initial system prompt $p_0$, training dataset $\mathcal{D}_{train}$, \#iterations (search-depth) $T$, \#prompts (breadth) $B=10$
\State Initialize $S_0 = \{p_0\}$, $i = 0$
\While{$i < T$}
    \State $i \gets i + 1$; $S_i \gets \emptyset$; $\text{max\_exp\_factor} \gets B/|S_{i-1}|$
    \For{each prompt $p_j^{i-1} \in S_{i-1}$}
        \State Evaluate $p_j^{i-1}$ on $\mathcal{D}_{train}$, and identify failure cases
        \State Generate feedback (using optimizer) \Comment{Section~\ref{sec:protegi-gradient-template}}
        \State Use LLM (optimizer) to generate prompts incorporating feedback $\rightarrow S_{ij}$ \Comment{Section~\ref{asec:protegi-prompt-gen-using-gradients}}
        \State Use MC-based sampling to generate 2 semantically similar prompts $\rightarrow S_{ij}^{mc}$ \Comment{Section~\ref{asec:protegi-semantic-prompt-gen}}
        \State $S_{ij} \gets S_{ij} \cup S_{ij}^{mc}$; 
        \If {$|S_{ij}| > \text{max\_exp\_factor}$} 
            \State  $S_{ij} \gets random.sample(S_{ij}, \text{max\_exp\_factor})$
        \EndIf
        \State $S_i \gets S_i \cup S_{ij}$
    \EndFor
    \State Compute macro F1 for each prompt in $S_i$ (using scorer) as per Successive Halving \cite{audibert2010best} with $\mathcal{D}_{mini} = \mathcal{D}_{train}/2$
    \State $S_i \leftarrow $ top 3 prompts based on macro F1
\EndWhile
\State \textbf{Output:} Best prompt $p_T$ over all iterations based on macro F1
\end{algorithmic}
\end{algorithm}

Exemplar selection is a unique challenge, and we intentionally refrained from incorporating examples in the prompt templates of the methods evaluated. 
Consequently, we exclude in-context learning (ICL) methods from our comparison to ensure fairness and plan to explore them in future work.


%% file: results-discussions.tex
\section{Results}

All methods begin with the same initial system prompt template~\ref{sec:base-system-prompt}, which does not include label definitions.
We observe that the final prompts for all methods except APE contain label definitions, which naturally emerge through the iterative optimization process. The final prompts selected for the Lisbon destination are available in Section~\ref{asec:best-prompts-all-methods}. For each method, the prompt with the highest macro F1 across 10 iterations is selected for test set evaluation.

\subsection{Cost vs. Performance Trade-off}
The cost–performance trade-off across five prompting methods, including CoT, evaluated over the top ten destinations is summarized in Figure~\ref{fig:costvsperformance}.
The following key insights emerge from the analysis:
\begin{enumerate}[wide, labelwidth=!, labelindent=0pt]
\item Cost is not linearly correlated with performance—higher cost does not always guarantee better results.
\item \textbf{ProTeGi} achieves the highest overall performance while remaining more cost-effective than both \textbf{OPRO} and \textbf{APE-OPRO}, although it incurs higher execution time as noted in~\cite{protegi}.
\item \textbf{APE-OPRO} matches \textbf{OPRO’s} performance with significantly lower cost, demonstrating the value of combining prompt initialization and iterative refinement.
\end{enumerate}
A comprehensive performance comparison of all methods across  10 destinations is provided in Appendix~\ref{asec:10-dest-perf}.

\subsection{APE-OPRO vs OPRO}
\begin{figure}
    \centering
    \includegraphics[width=0.75\linewidth]{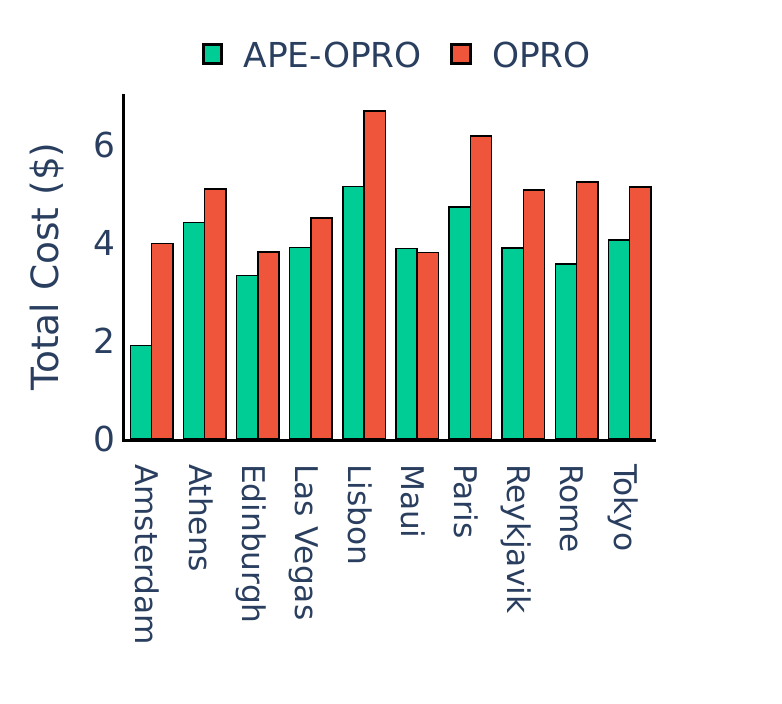}
    \caption{Total cost comparison between APE-OPRO and OPRO across 10 destinations. Each bar reflects the aggregated cost over 10 optimization iterations with 10 candidate prompts per iteration}
    \label{fig:ape-opro-vs-opro-cost-10dest}
\end{figure}

This section analyzes why APE-OPRO offers better cost-efficiency while maintaining similar performance compared to OPRO.
As shown in Figure~\ref{fig:ape-opro-vs-opro-cost-10dest}, OPRO consistently incurs higher costs across all destinations, with Amsterdam showing the most significant difference, nearly twice the cost of APE-OPRO. On average,
OPRO incurs ~18\% higher costs under our standard run configuration.

This disparity in cost stems from their initialization strategies. APE-OPRO begins with APE generated prompts, which are semantically similar variants of the initial system prompt~\ref{sec:base-system-prompt}. These prompts are scored using macro F1, and the top 3 are fed into the metaprompt to produce structured prompts with label definitions, continuing this cycle until a fixed number of iterations. In contrast, OPRO starts with the metaprompt directly, without prior prompts or scores. As a result, it creates long and detailed prompts from the beginning, which then lead to even longer prompts in later steps.

\begin{figure}
    \centering
    \includegraphics[width=0.75\linewidth]{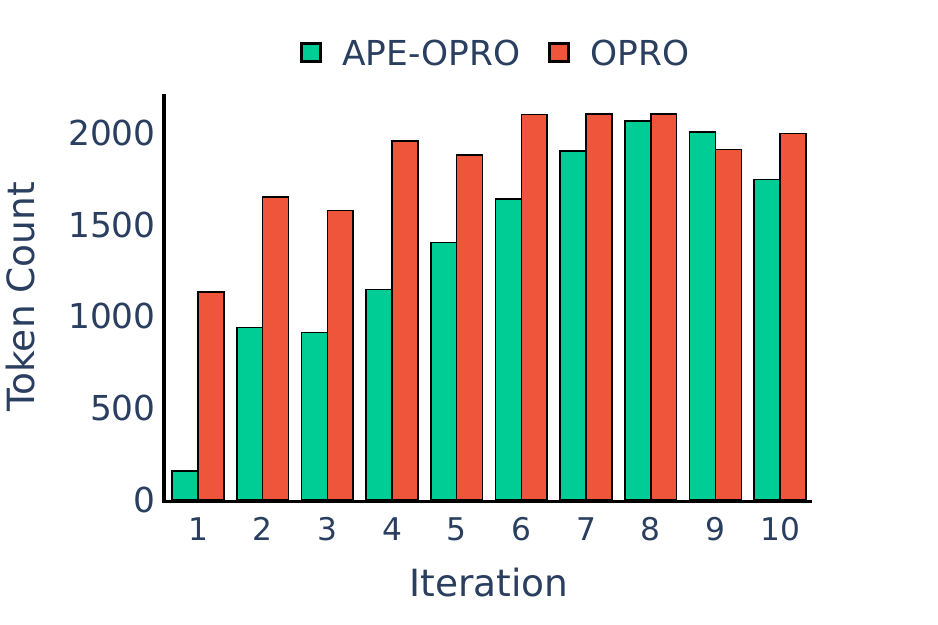}
    \caption{Distribution of token count for the best prompt saved in each optimization iteration for APE-OPRO and OPRO in Amsterdam}
    \label{fig:ape-opro-vs-opro-token-count-periter-amst}
\end{figure}

We illustrate this distinction in Figure~\ref{fig:ape-opro-vs-opro-token-count-periter-amst} using Amsterdam as a representative case, where the divergence is particularly pronounced and serves as a clear example to contextualize the observed behavior.
APE-OPRO starts with a lower token count in the first iteration, due to its APE initialization, and increases gradually. 
OPRO, by contrast, exhibits a steep early rise in token count, plateauing around iteration 6. For illustration, we include the best prompts from both methods at iterations 1 and 4 in Appendix~\ref{asec:apeopro-vs-opro}. Iteration 1 highlights initialization differences, while iteration 4 shows the significant difference in token counts between two methods.

This difference in approach is analogous to gradient descent, where OPRO attempts a large optimization jump in the first iteration, 
and potentially diverges from a cost-efficient solution.
In contrast, APE-OPRO follows a more gradual, stepwise refinement process starting from a simple prompt and improving it iteratively, leading to more efficient prompt evolution at reduced cost, while achieving similar performance.

\input{res-variance-analysis}
\subsection{Convergence Analysis}
\label{sec:convergence-analysis}
\begin{figure}
    \centering
    \includegraphics[width=1.15\linewidth, keepaspectratio]{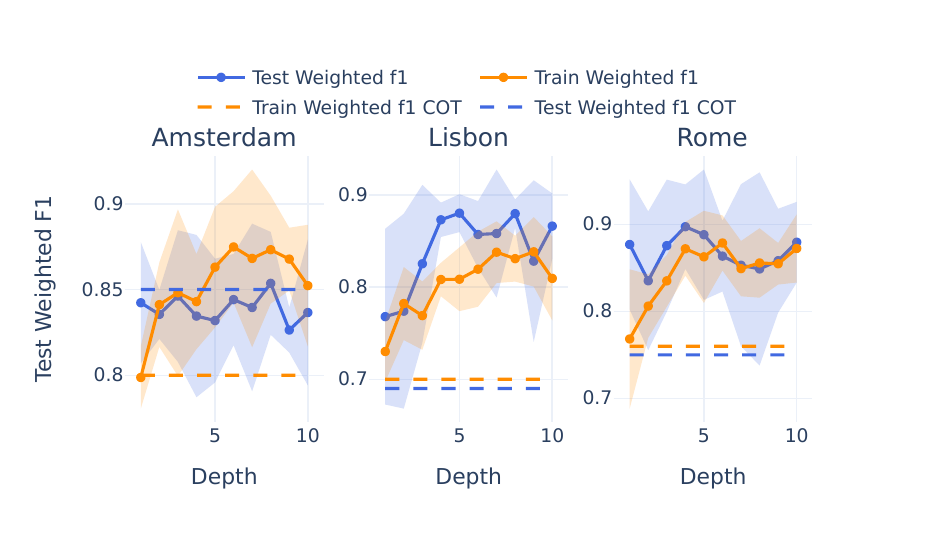}
    \caption{Convergence analysis for ProTegi on 3 key destinations}
    \label{fig:ProTegi_weightedf1_3_dest}
\end{figure}

Convergence analysis evaluates model stability and generalization as reasoning depth increases, helping determine whether further depth yields meaningful gains or if performance has plateaued.
We evaluate this by tracking the mean and variance of weighted F1 scores on both training and test sets across increasing depths, averaging results over five independent runs per method and across three destinations. Figure~\ref{fig:ProTegi_weightedf1_3_dest} shows these trends for ProTeGi, while convergence plots for other methods appear in Section~\ref{asec:convergence_analysis}. Several notable patterns emerge:
\begin{enumerate}[wide, labelwidth=!, labelindent=0pt, nosep]
\item ProTeGi consistently shows strong convergence and generalization across all three destinations, with a narrow train-test gap and performance largely stabilizing by iteration 5–6.
\item For Amsterdam, CoT performs well even at shallow depths, leaving limited room for further gains. Most models plateau early.
\item For Rome, all methods demonstrate continued performance improvement up to depth 10, suggesting potential for further gains with increasing depth.
\item For Lisbon, both ProTeGi and APE-OPRO show rapid gains from the start and converge by depth 5–6, while OPRO achieves most of its improvement in the first iteration and plateaus by depth 1.
\item Interestingly, all 4 methods often show lower weighted F1 scores on the training set than on the test set. This counter-intuitive pattern stems from our sampling strategy, limiting training to at most four examples per label, thus making training metrics highly sensitive to individual errors. In contrast, the larger test set yields more stable and representative performance scores.
\end{enumerate}
These findings collectively highlight the efficiency and generalization strength of ProTegi, while also illustrating destination-specific dynamics in convergence behavior. 

\subsection{Ablating on Depth}
\label{sec:ablating_depth}
For all methods except CoT, with no principled stopping criterion, the number of optimization iterations (or depth) is a key hyperparameter. As such, the optimal depth might vary across methods and datasets. 
We evaluated each method at multiple depths to assess its impact on performance and cost, with ProTeGi's \texttt{search-depth} serving as its analogous control.

\begin{figure}
    \centering
    \includegraphics[width=1.05\linewidth,]{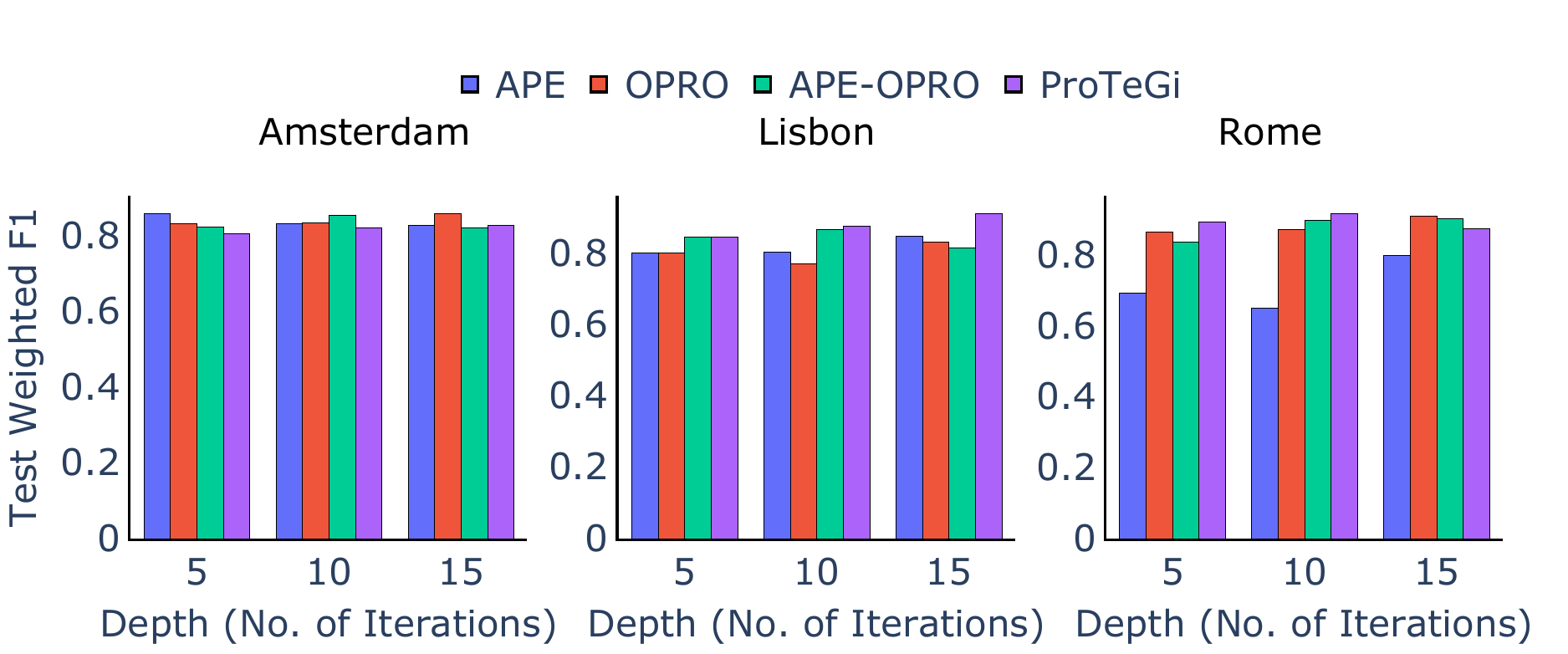}
    \caption{Change in Performance for depth or maximum number of iterations for each method across destinations}
    \label{fig:depth_test_wf1}
\end{figure}

Figure \ref{fig:depth_test_wf1} reports the test set weighted F1 scores 
for three destinations across different iteration counts.
Due to low variance observed in performance metrics \ref{sec:variance-analysis}, we present results from a single run per configuration.

As expected, increasing iteration count significantly raises costs. Moving from 5 to 10 and 15 iterations typically resulted in a 2x and 3x cost increase, respectively. ProTeGi showed a slightly steeper rise, averaging 2.3–2.6x and 3.3–3.6x (see Appendix, Figure \ref{fig:depth_test_wcost_wf1}). 
While scorer output tokens increase linearly with depth, ProTeGi's optimizer input/output and scorer input tokens grow super-linearly, likely due to more detailed feedback generating longer prompts. 
However, this increased specificity did not consistently yield better performance. For ProTeGi, test weighted F1 scores sometimes plateaued or declined with increased iteration depth (from 5 to 10 and 10 to 15), consistent with trends in the original \cite{protegi} paper. Similar diminishing returns were observed in other methods, suggesting that smaller depths may suffice for effective prompt optimization.

\subsection{Ablating on Breadth}
\label{sec:ablating_breadth}
Another key hyperparameter across all evaluated methods is the breadth—the number of prompts generated and evaluated per iteration. This controls the size of the candidate pool assessed during each round to select the top ($k = 3$) prompts for the next iteration. In \protegi, this is implemented during the selection phase (successive halving), where a fixed number of prompts are forwarded if the post-expansion pool exceeds the specified breadth.

\begin{figure}
    \centering
    \includegraphics[width=1.05\linewidth]{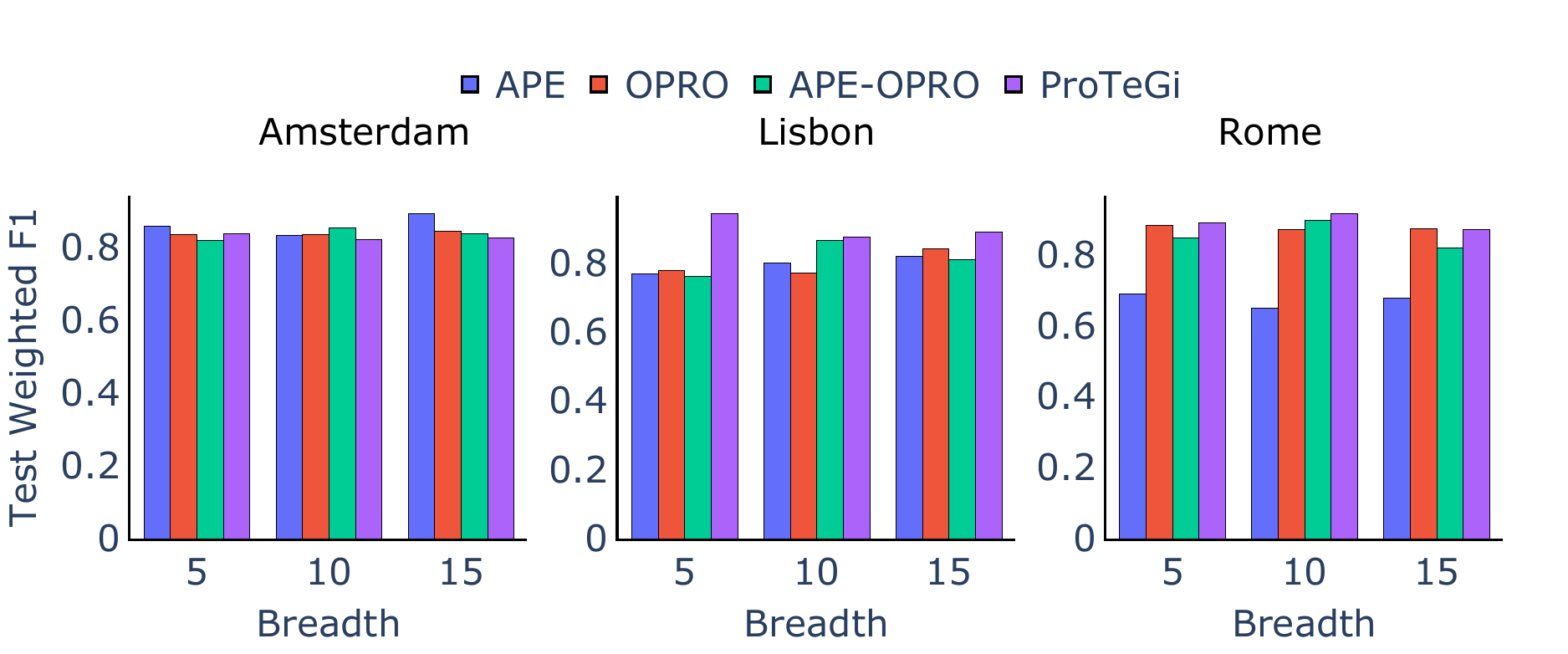}
    \caption{Change in Performance for breadth or maximum number of prompts in each iteration for each method across destinations}
    \label{fig:breadth_test_wf1}
\end{figure}

Figure \ref{fig:breadth_test_wf1} shows the impact of varying breadth (5, 10, 15) on test weighted F1 scores 
across three destinations.
As expected, increasing breadth resulted in a roughly 2x and 3x cost increases for most methods when moving from 5 to 10 and 15 prompts, respectively, ProTeGi showed a more subdued cost increase (refer Appendix, Figure \ref{fig:breadth_test_wcost_wf1}). 
This is because breadth in ProTeGi mainly impacts the scorer model during selection, with prompt generation in the expansion phase governed by separate, fixed parameters (e.g., number of feedbacks per error group, gradients per feedback, and prompts per gradient) that were not varied in this study. 

Similar to depth, increasing breadth yields marginal or negative gains in test weighted F1 scores, indicating diminishing returns. This suggests that smaller breadth values may suffice for effective prompt optimization across all methods.

\subsection{Sensitivity to Label List Formatting in Prompt Templates}
\begin{figure}[t]
    \centering
    \includegraphics[height=3.9cm, keepaspectratio, width=\linewidth]{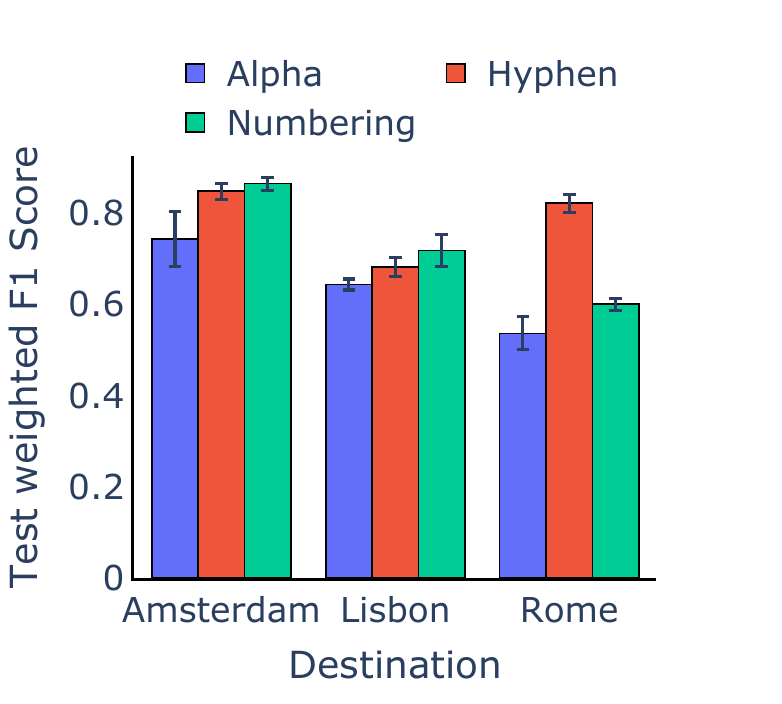}
    \caption{Effect of Label Formatting Styles such as hyphens (-), numbers ($1., 2., 3., \cdots $), alphabets ($a., b., c., \cdots $) on APE's Performance}
    \label{fig:labelordering}
\end{figure}

To evaluate the impact of label formatting on prompt optimization performance while keeping label order fixed, we investigated how different styles of presenting label lists such as hyphenation (- Label), numeric prefixes (1. Label), and alphabetical prefixes (a. Label) affect model behavior within the initial system prompt~\ref{sec:base-system-prompt}. The corresponding prompt templates for these formatting styles are provided in Appendix~\ref{asec:label-ordering}. This experiment was conducted across all four methods and we explicitly stated in the prompt: {\em Treat all labels as equally likely and independent of their position in the list}. Despite this explicit guidance, we observed that APE was notably sensitive to changes in label formatting, particularly when labels were numerically or alphabetically indexed. In contrast, the other methods remained largely unaffected (see Appendix Section~\ref{asec:label-ordering}).
To quantify this effect, we ran five independent trials of APE on the three key destinations, reporting the test weighted F1 scores and their variance in Figure~\ref{fig:labelordering}.

A likely reason for this sensitivity is that APE generates final prompts without including label definitions, relying only on the label names. In contrast, other methods generate detailed label definitions, which may help reduce the impact of formatting effects. 

%% file: res-variance-analysis.tex
\subsection{Variance Analysis}
\label{sec:variance-analysis}
\begin{figure}
    \centering
    \includegraphics[width=0.7\linewidth]
    {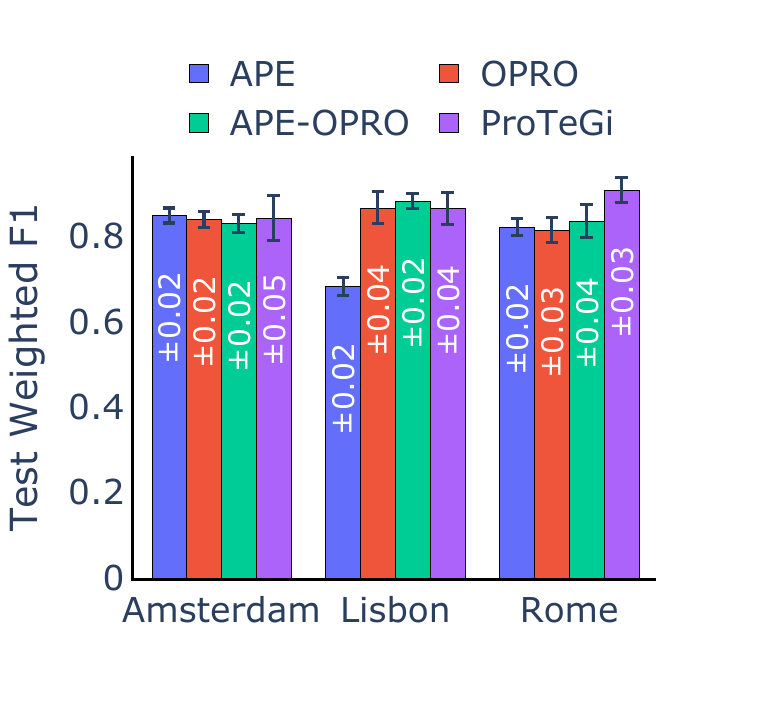}
    \caption{Evaluation of method generalizability across diverse destination datasets, showing the average weighted F1 score and standard deviation over five independent test trials per method}
    \label{fig:test-var-wf1}
\end{figure}

Given the inherent stochasticity of LLMs, 
quantifying the variance in model performance is crucial for assessing generalizability.
Figure~\ref{fig:test-var-wf1} presents the average test weighted F1 scores computed over five independent runs for each method across three distinct destination datasets. Each bar in the figure is annotated with the corresponding standard deviation, and error bars are included to visualize variability.

Our analysis reveals that model performance exhibits noticeable variation across destinations, underscoring the influence of dataset-specific characteristics on the efficacy of each method. 
Methods such as \ape, \opro, and \apeopro consistently exhibit lower variance across all destinations, suggesting more stable behavior under repeated evaluations.
The average standard deviation in performance across the three destinations is approximately $0.04$ for \protegi, followed by $0.03$ for \opro, $0.026$ for \apeopro, and $0.02$ for \ape.
We hypothesize that one contributing factor to the slightly higher variance observed in \protegi may be the limited size of the sampled training dataset used in our implementation, where $\mathcal{D}_{mini} = \mathcal{D}_{train} / 2$, as opposed to the larger dataset size $\mathcal{D}_{mini} = 64$ recommended by the original authors \cite{protegi}.

%% file: conclusion.tex
\section{Conclusion and Future work}
In this work, we addressed a critical gap by providing the first comprehensive evaluation of APO methods for 
real-world, commercially relevant multiclass classification task.
We systematically benchmarked prominent gradient-free (APE, OPRO) and gradient-based (ProTeGi) approaches on our proprietary dataset and introduced APE-OPRO, a novel hybrid framework.
Our empirical analysis revealed key performance-cost trade-offs relevant to practical implementation.
While ProTeGi demonstrated superior API cost-efficiency and performance, 
it required longer execution times.
APE-OPRO offered a compelling balance by matching OPRO's performance, outperforming APE, and significantly reducing cost compared to OPRO. 
These findings provide valuable insights for practitioners, allowing them to choose between optimizing for API cost (using ProTeGi) or balancing speed and performance (using APE-OPRO), based on their specific computational and time limitations. This versatility underscores APO's effectiveness for complex commercial applications.

Moving forward, our study highlights several promising avenues for future research in APO. 
Key directions include developing methods for more structured and fine-grained control over prompt updates~\cite{GPO}, as current techniques offer limited granularity. 
Expanding the scope of APO is also critical, particularly to address multi label tasks and to extend optimization techniques to vision and multimodal domains, which remain largely underexplored. 
Furthermore, investigating the impact of incorporating persona-based prompting (for example, using destination-specific personas) within the optimization process presents another interesting direction. Finally, exploring the use of different models as scorers and optimizers may offer new insights into performance–cost trade-offs.

%% file: appendix-all.tex
\clearpage
\onecolumn
\section{Appendix}
\input{appendix/performance-10-dest}
\clearpage
\input{appendix/metrics}

\clearpage
\subsection{Sensitivity to Label List Formatting in Prompt Templates}
\label{asec:label-ordering}
\input{appendix/label-ordering}

\clearpage
\subsection{APE OPRO vs OPRO}
\label{asec:apeopro-vs-opro}
\input{appendix/ape-opro-best-prompt-iter1-amsterdam}
\input{appendix/opro-best-prompt-iter1-amsterdam}
\input{appendix/ape-opro-best-prompt-iter4-amsterdam}
\input{appendix/opro-best-prompt-iter4-amsterdam}

\clearpage
\subsection{Prompt Templates}
\label{asec:prompt_templates}
\input{appendix/base-system-prompt-template}
\input{appendix/generate-similar-prompt-template}
\input{appendix/opro-metaprompt-template}
\input{appendix/protegi-gradient-template}

\clearpage
\subsection{Best Prompts}
\label{asec:best-prompts-all-methods}
\input{appendix/ape-best-prompt-lisbon}
\input{appendix/opro-best-prompt-lisbon}
\input{appendix/ape-opro-best-prompt-lisbon}
\input{appendix/protegi-best-prompt-lisbon}

\clearpage
\subsection{Variance Analysis}
\label{asec:variance_analysis}
\begin{figure}[h]
    \centering
    \includegraphics[width=0.5\linewidth]{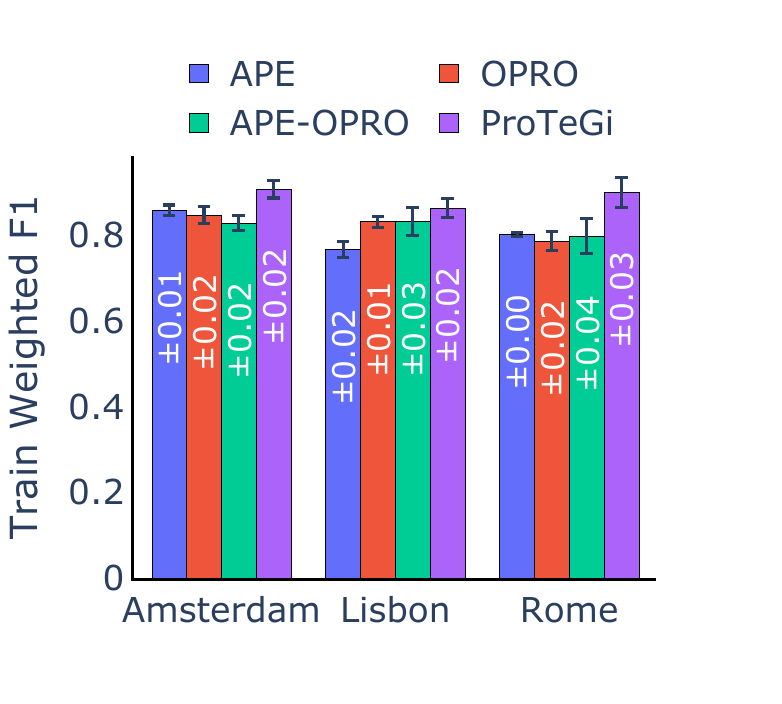}
    \caption{Variance in the train weighted F1 score for all methods, across three different destinations for five independent trials.}
    \label{fig:var_train_wf1}
\end{figure}

Figure \ref{fig:var_train_wf1} presents the variance in the weighted F1 scores on the training dataset. As anticipated, the standard deviation on the training data is relatively low across most methods when compared to the test data. This aligns with expectations, given that models typically fit better on training data. Notably, APE-OPRO exhibits a slightly higher standard deviation on the training set, averaging $\pm0.03$, in contrast to $\pm0.0266$ observed on the test set.

\begin{figure}[h] 
    \centering
    \begin{minipage}[b]{0.49\linewidth} 
        \centering
        \includegraphics[width=\linewidth]{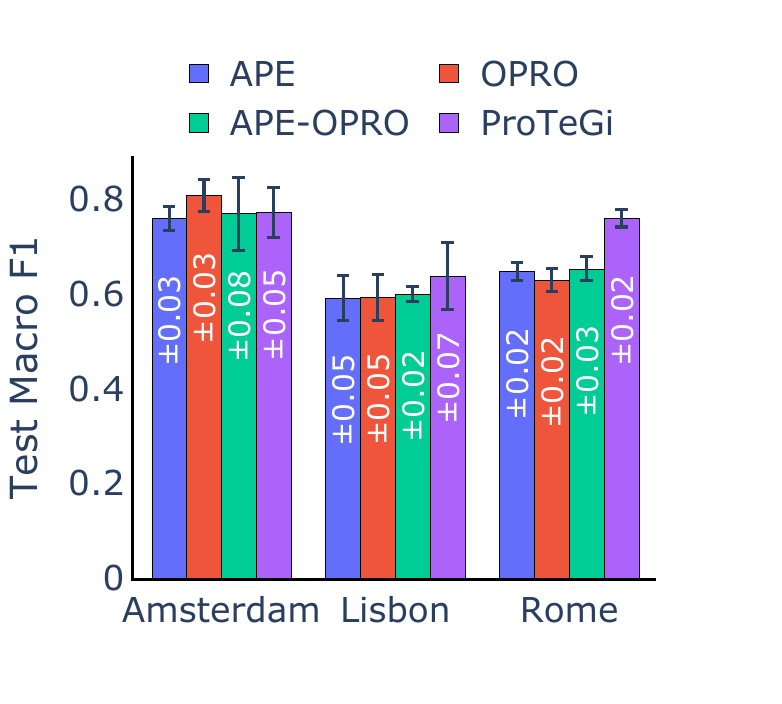}
    \end{minipage}
    \hfill 
    \begin{minipage}[b]{0.49\linewidth} 
        \centering
        \includegraphics[width=\linewidth]{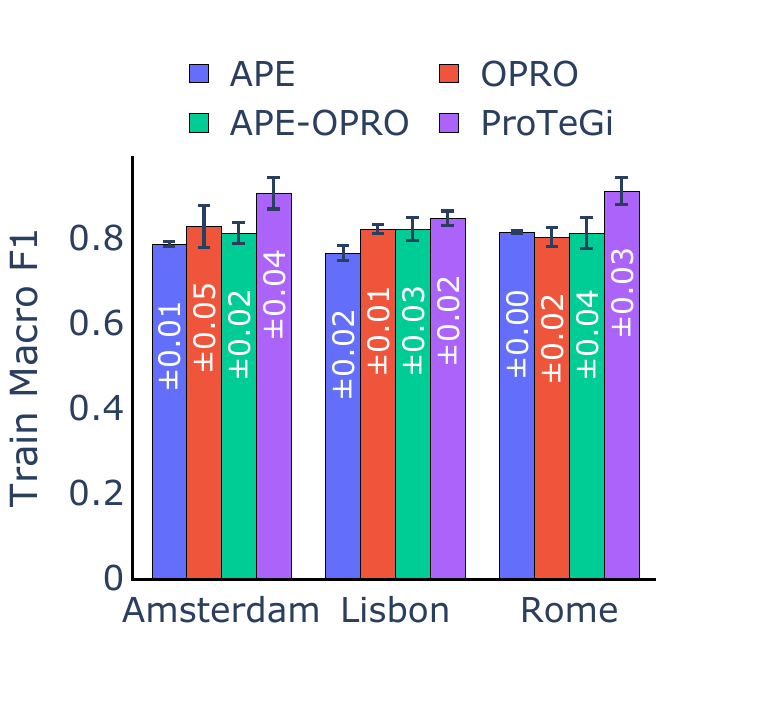}
    \end{minipage}
    \caption{Evaluation of method generalizability across diverse destination datasets, showing the avg. weighted F1 score and standard deviation over five independent test and train trials per method.} 
    \label{fig:combined_var_figures} 
\end{figure}

Figure \ref{fig:combined_var_figures} shows the standard deviation of the macro F1 score on both the test and training datasets. As briefly discussed in Section~\ref{sec:evaluation_metric}, macro F1 is inherently sensitive to class imbalance, particularly due to the label distribution in the test set. This sensitivity contributes to both the lower overall macro F1 scores and the higher observed standard deviation on the test data.


\clearpage
\subsection{Ablating on Depth}
\label{asec:ablating_depth}
\begin{figure*}[h]
    \centering
    \includegraphics[scale=0.5]{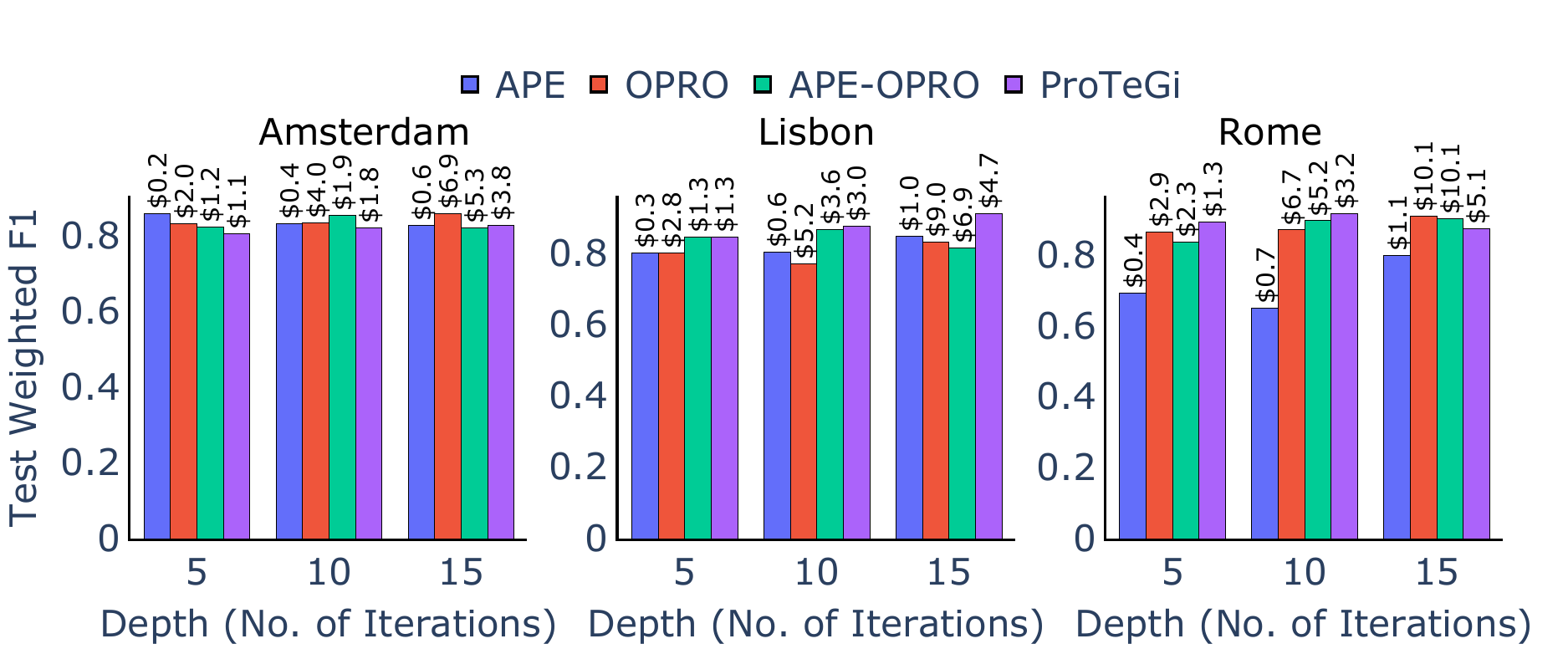}
    \caption{Change in Performance for depth or maximum number of iterations for each method across destinations along with the (API) cost incurred for each run of the respective algorithm.}
    \label{fig:depth_test_wcost_wf1}
\end{figure*}

Figure \ref{fig:depth_test_wcost_wf1} presents the test weighted F1 scores across three key destinations for all four methods, along with the corresponding costs incurred as depth increases. 
This plot is similar to the plot in section \ref{sec:ablating_depth} with cost information included for each method for each destination.

\begin{figure*}[h]
    \centering
    \includegraphics[scale=0.5]{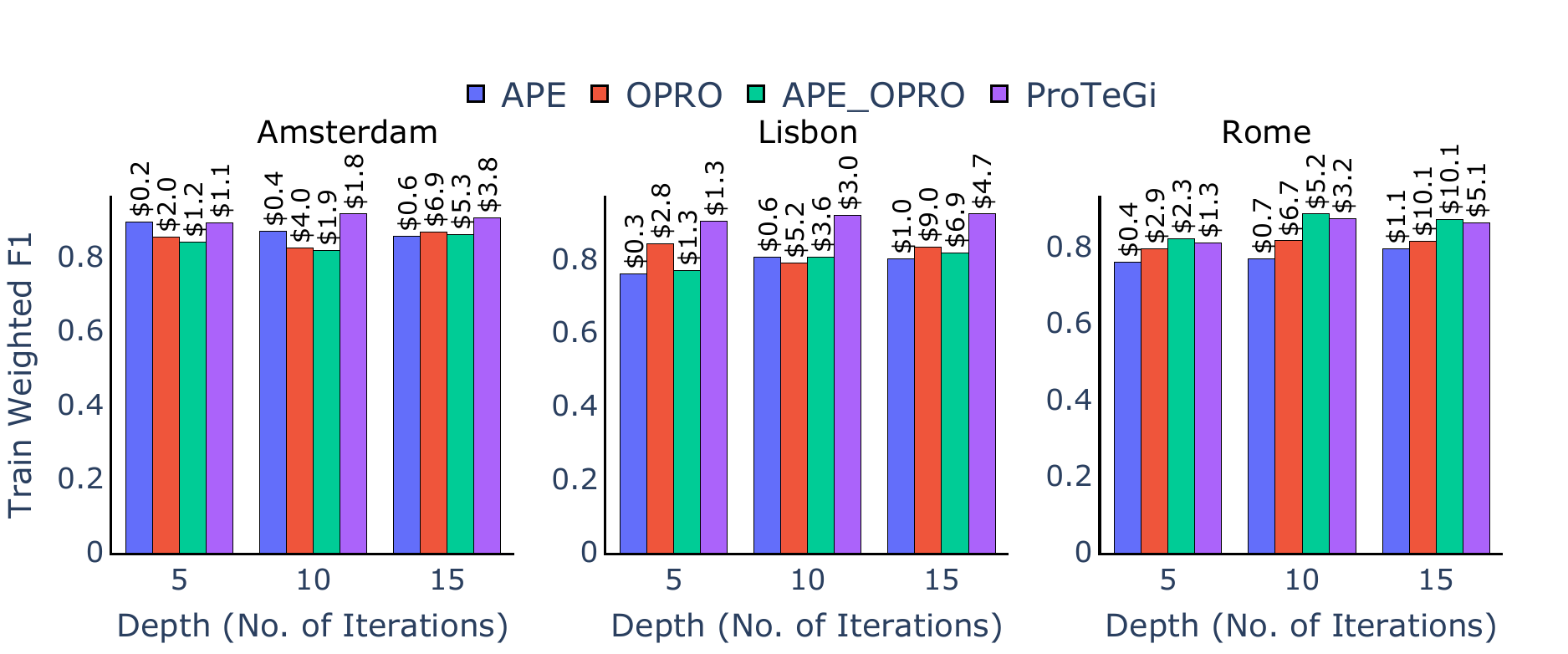}
    \caption{Caption}
    \label{fig:depth_train_wf1}
\end{figure*}

Figure~\ref{fig:depth_train_wf1} presents the training weighted F1 scores across three key destinations for all four methods, along with the associated costs as depth increases. Note that the cost values are identical to those reported in Figure~\ref{fig:depth_test_wf1}. Across all methods, diminishing returns in training weighted F1 were observed with increasing depth, suggesting that smaller depths may be sufficient for effective prompt optimization which is consistent with the pattern seen in the test weighted F1 results.

\clearpage
\subsection{Ablating on Breadth}
\begin{figure*}[h]
    \centering
    \includegraphics[scale=0.5]{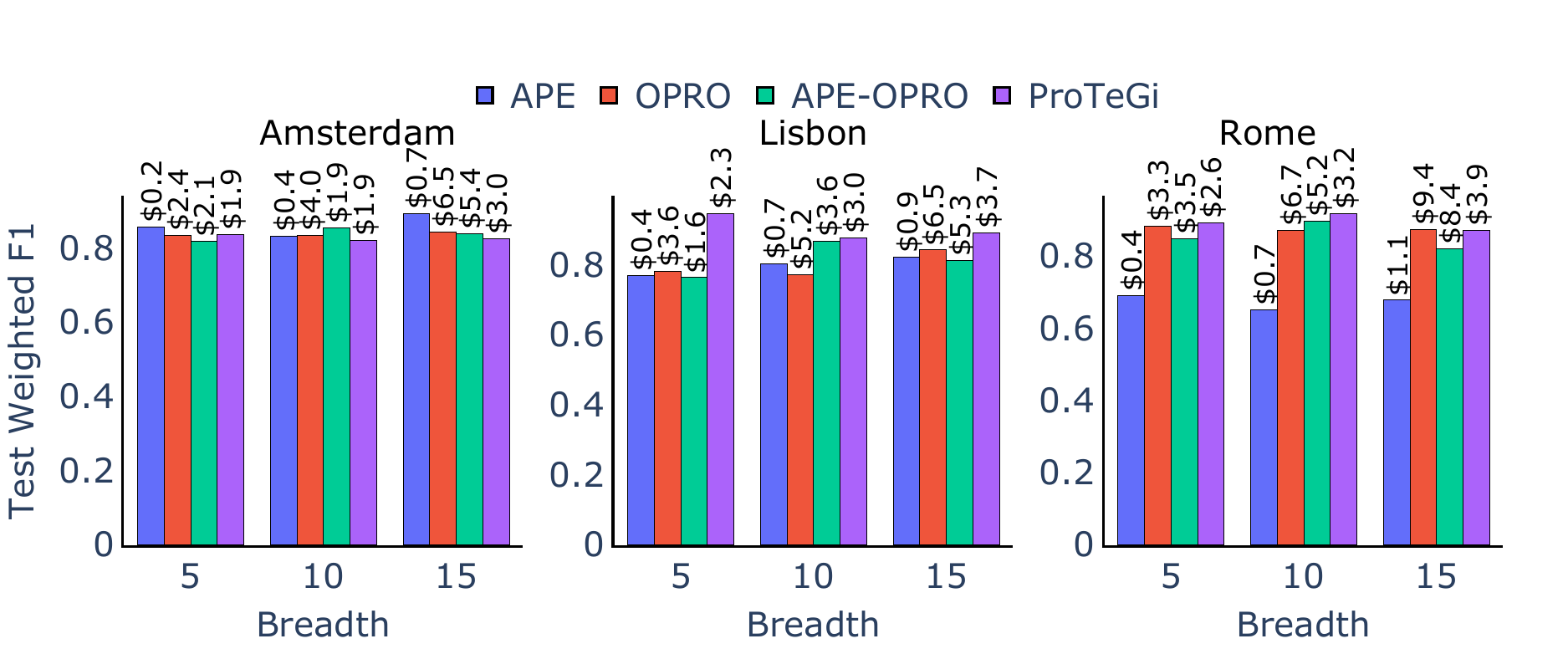}
    \caption{Change in Performance for breadth or maximum number of prompts in each iteration for each method across destinations}
    \label{fig:breadth_test_wcost_wf1}
\end{figure*}

Figure \ref{fig:breadth_test_wcost_wf1} presents the test weighted F1 score and associated cost as breadth increases. 
This plot is similar to the plot in section \ref{sec:ablating_breadth} with cost information included for each method for each destination.

\begin{figure*}[h]
    \centering
    \includegraphics[scale=0.5]{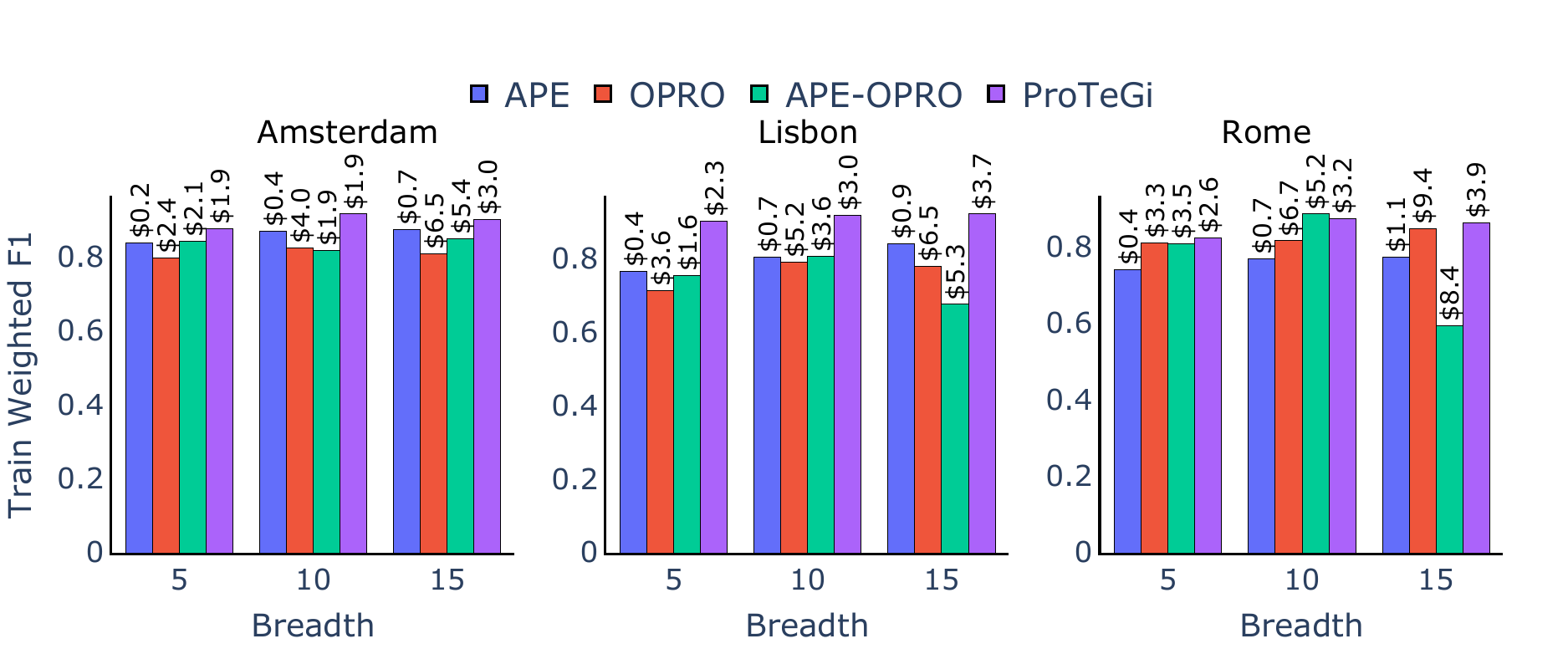}
    \caption{Caption}
    \label{fig:breadth_train_wf1}
\end{figure*}

Figure~\ref{fig:breadth_train_wf1} displays the training weighted F1 score and corresponding cost as breadth increases. The observed diminishing or negative returns in F1 performance suggest that lower breadth values are likely adequate for effective prompt optimization.


\clearpage
\subsection{Convergence Analysis}
\label{asec:convergence_analysis}
The following graphs present the convergence analysis of APE, OPRO, and APE-OPRO for 3 key destinations and these are similar to the plot in section~\ref{sec:convergence-analysis}
\begin{figure}[h]
    \centering
    \includegraphics[width=\linewidth,
    keepaspectratio]{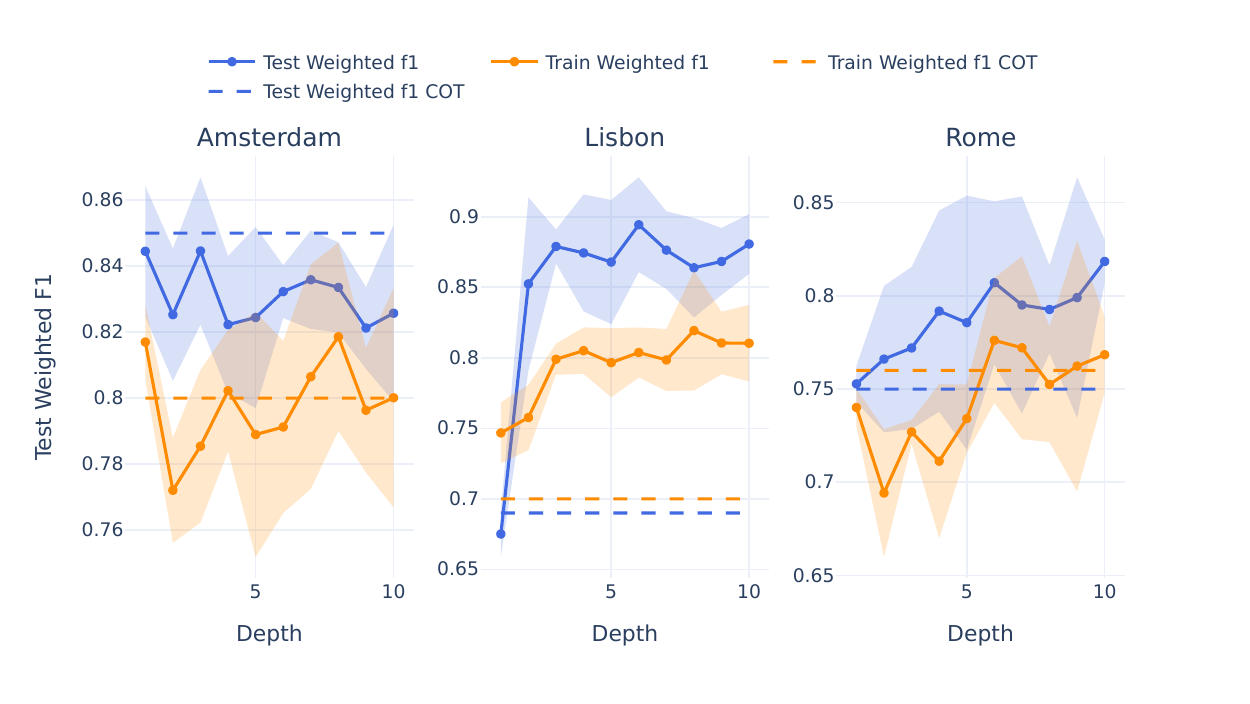}
    \caption{Convergence analysis for APE-OPRO on 3 key destinations}
    \label{fig:APE_OPRO_weightedf1_3_dest}
\end{figure}

\begin{figure}[h]
    \centering
    \includegraphics[width=\linewidth,
    keepaspectratio]{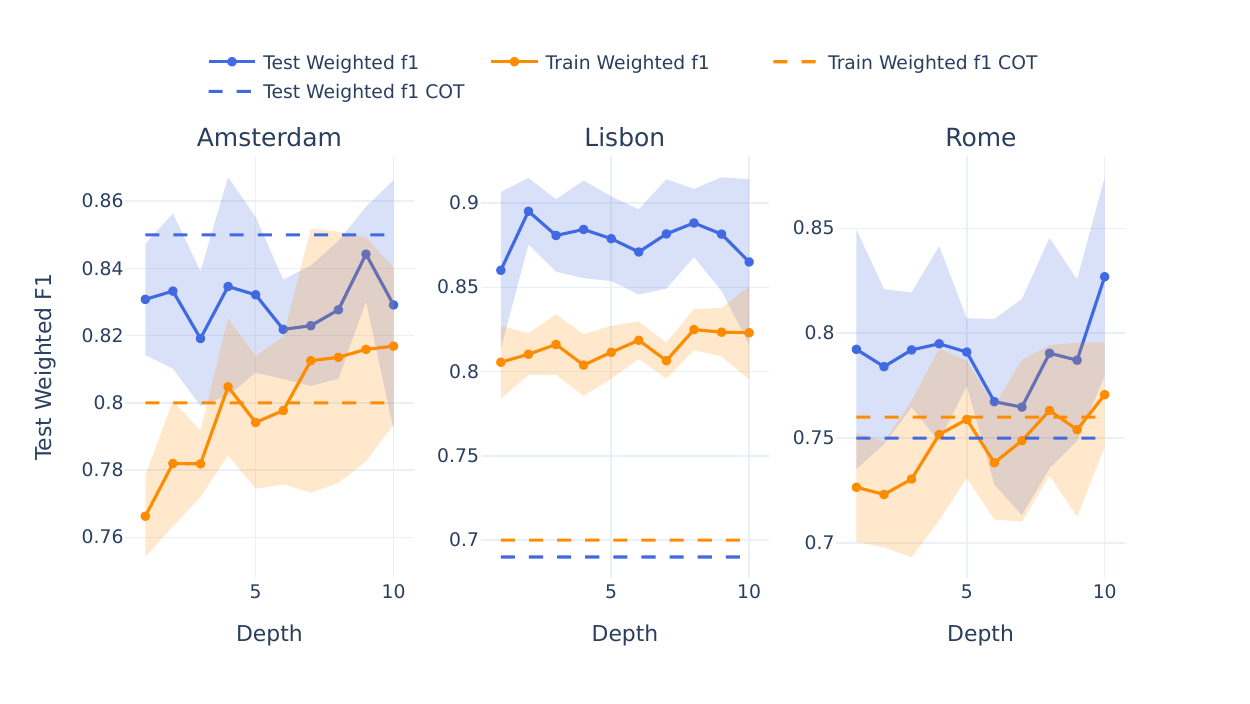}
    \caption{Convergence analysis for OPRO on 3 key destinations}
    \label{fig:OPRO_weightedf1_3_dest}
\end{figure}

\begin{figure}[h]
    \centering
    \includegraphics[width=\linewidth,
    keepaspectratio]{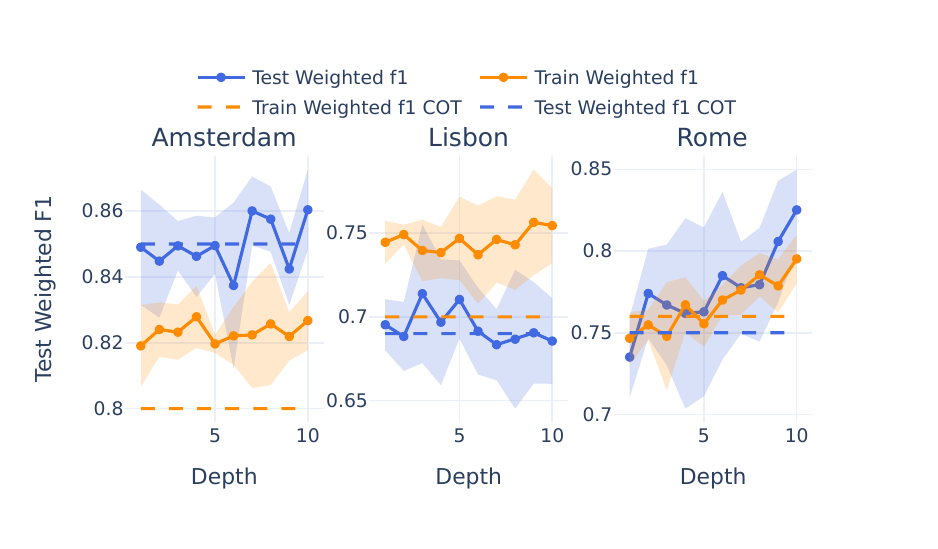}
    \caption{Convergence analysis for APE for 3 key destinations}
    \label{fig:APE_weightedf1_3_dest}
\end{figure}

%% file: appendix/performance-10-dest.tex
\subsection{Train Test Performance on 10 destinations}
\label{asec:10-dest-perf}
\begin{figure*}[h]
    \centering
    \includegraphics[width=0.90\linewidth]{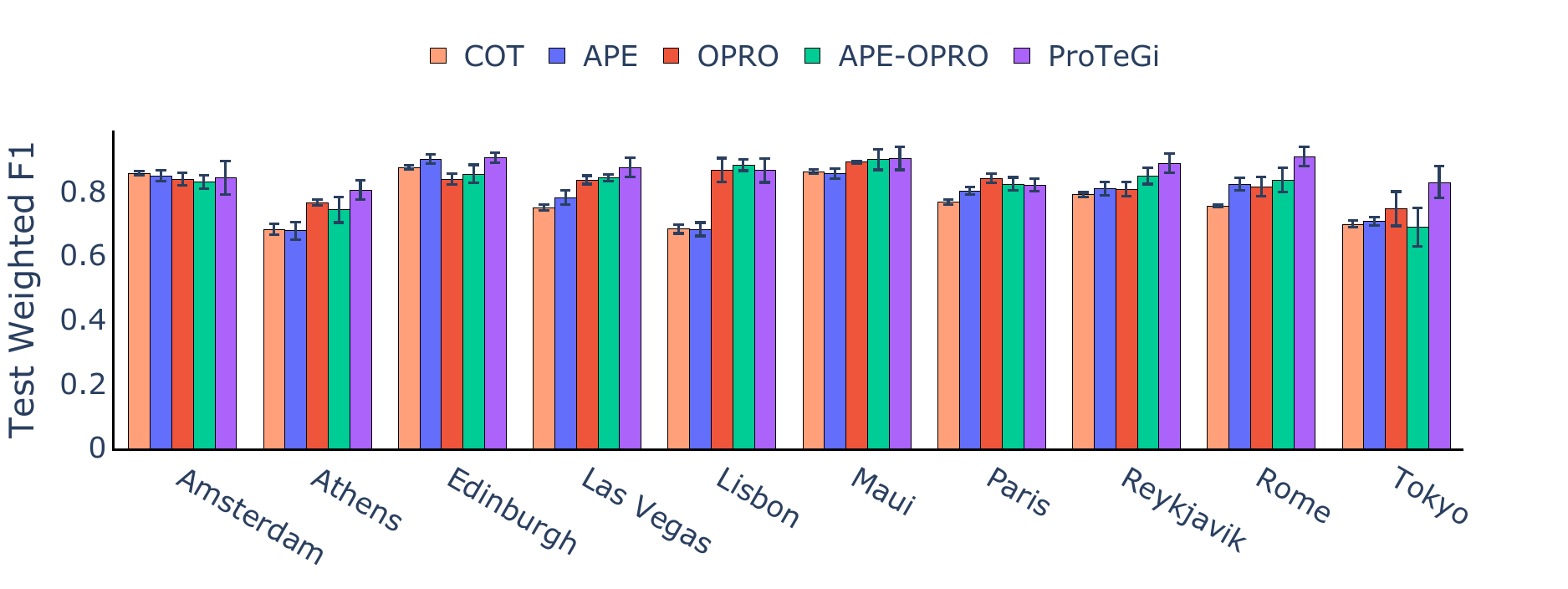}
    \caption{Test weighted F1 (averaged over 5 runs) on Top 10 destinations}
    \label{fig:test_weightedf1_10dest}
\end{figure*}

\begin{figure*}[h]
    \centering
    \includegraphics[width=0.90\linewidth]{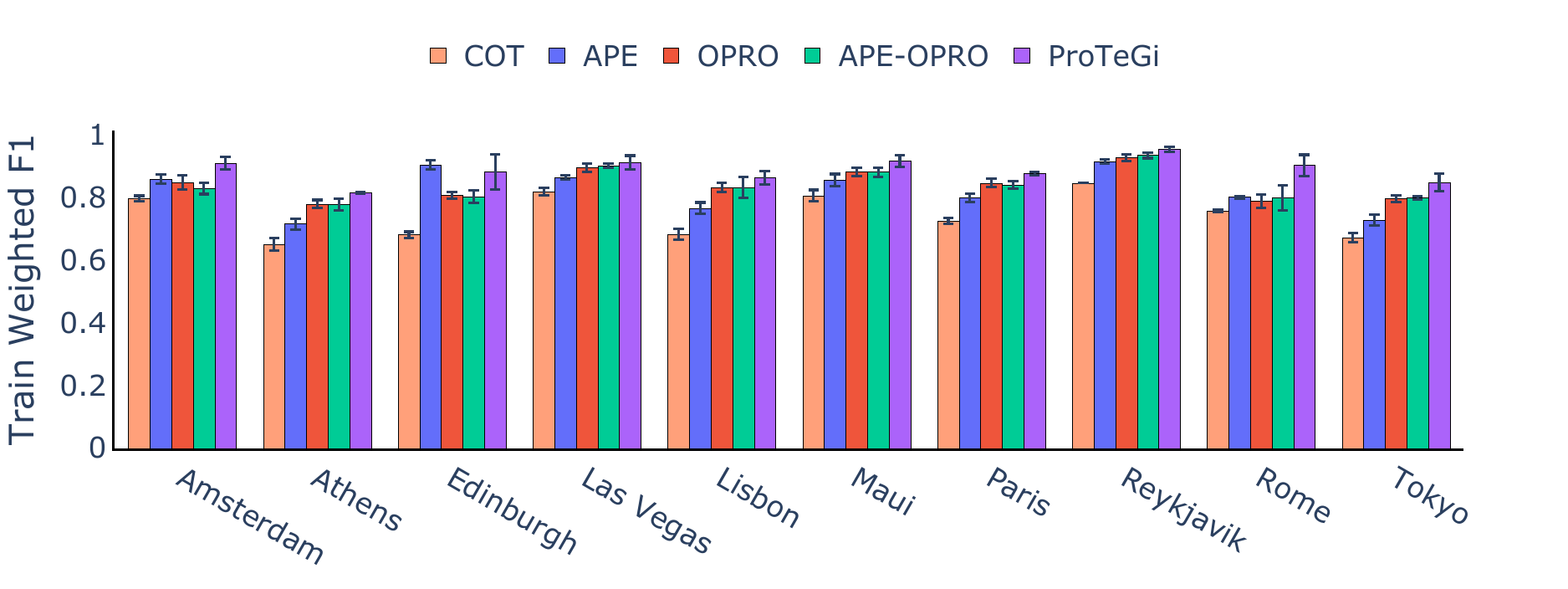}
    \caption{Train weighted F1 (averaged over 5 runs) on Top 10 destinations}
    \label{fig:train_weightedf1_10dest}
\end{figure*}

The Figures \ref{fig:test_weightedf1_10dest} and \ref{fig:train_weightedf1_10dest}  present the model’s performance on both test and train datasets, measured by the average weighted F1 score across 10 destinations. Results are averaged over five runs per destination, with standard deviation included to indicate variability in performance. 
Note that Figure \ref{fig:costvsperformance} is a summarized version of this plot showing the performance averaged over all the destinations for each method.
We observe that ProTeGi has either best or at par test performance for almost all the 10 destinations.  

%% file: appendix/metrics.tex
\subsection{Metrics}
\label{sec:metrics}
This task is a standard multiclass classification problem, where macro F1 and weighted F1 are common evaluation metrics. As shown in Figure~\ref{fig:label-dist-top10-dest}, the label distribution is highly imbalanced, with several classes appearing fewer than three times across major destinations. This skews the evaluation: macro F1, by weighting all classes equally, drops significantly compared to weighted F1, which accounts for frequency.

\begin{table}[ht]
    \centering
    \caption{Per-class F1 scores for Lisbon}
    \label{tab:per_class_f1_lisbon}
    \begin{tabular}{l c}
        \toprule
        \textbf{Label} & \textbf{F1 Score} \\
        \midrule
        City Sightseeing Cruises & 1.0000 \\
        Culinary Exploration Tour & 0.9756 \\
        Lisbon City Tours by Land & 0.9684 \\
        Sintra \& Coastal Highlights & 0.9268 \\
        Sintra Day Tours & 0.9143 \\
        Workshop or Class & 0.9091 \\
        Sacred \& Holy Sites & 0.8889 \\
        Coastal \& Scenic Tours & 0.8571 \\
        Nearby Cities and Town & 0.8571 \\
        Vineyards \& Wine Tasting & 0.8571 \\
        Transfer & 0.8000 \\
        Combo Experience Tour & 0.5714 \\
        Historical Tour & 0.5000 \\
        Bar Hopping & 0.0000 \\
        Iconic Sites Tour & 0.0000 \\
        Lisbon Hills & 0.0000 \\
        \bottomrule
    \end{tabular}
\end{table}

\begin{figure}[H]
    \centering
    \includegraphics[width=0.5\linewidth, keepaspectratio]{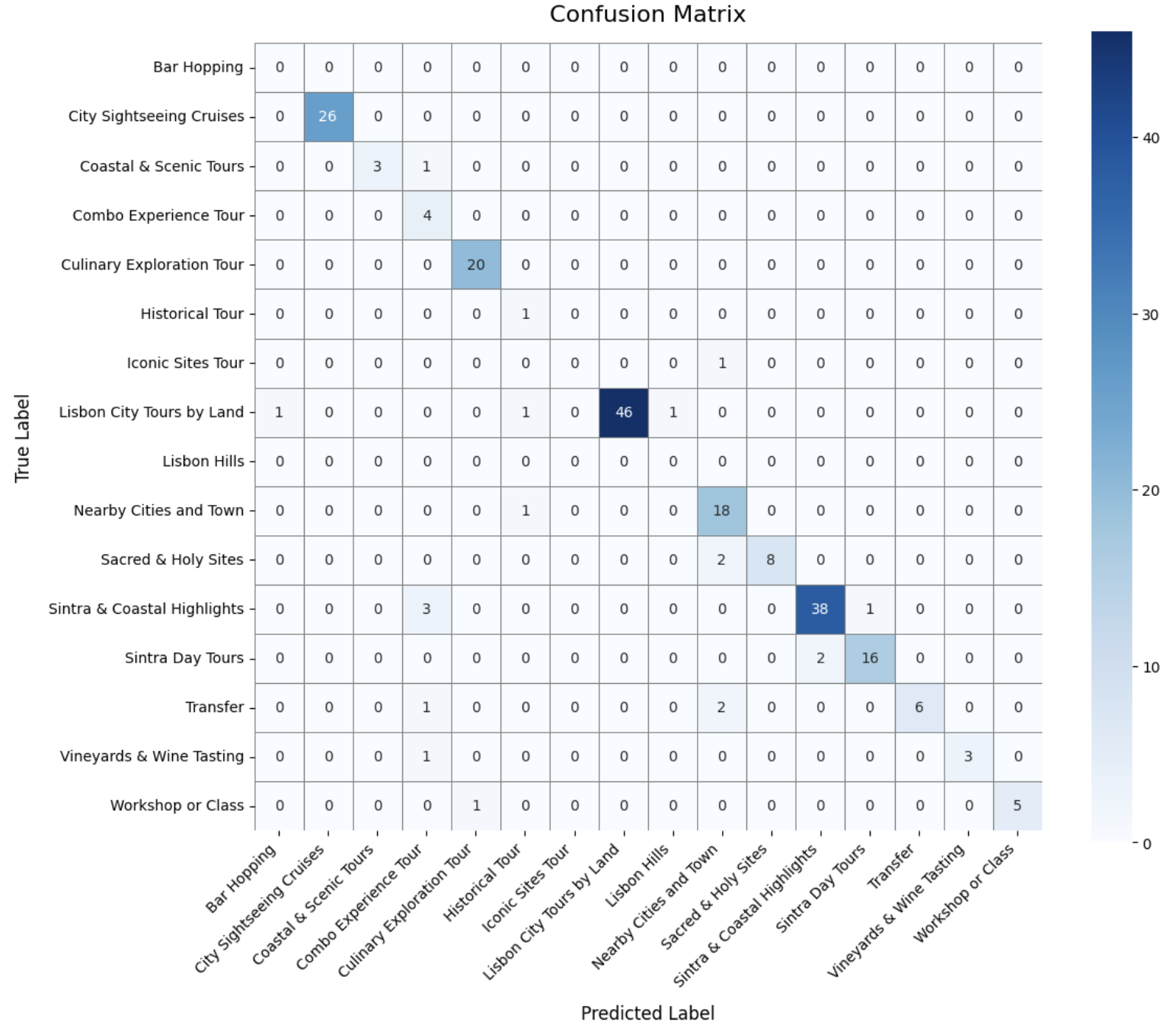}
    \caption{Confusion Matrix for Lisbon}
    \label{fig:confusionmatrix_lisbon}
\end{figure}

As shown in Figure~\ref{fig:confusionmatrix_lisbon}, in the Lisbon dataset, rare classes like Bar Hopping, Iconic Sites Tour, and Lisbon Hills each incur just a single error (false positive or false negative), yet receive an F1 score of zero due to macro F1’s sensitivity. While such edge cases are heavily penalized under macro F1, they have limited impact on real-world utility, where frequent classes dominate and labels serve primarily as metadata or filters on Viator listing pages.
Given this, we adopt weighted F1 as the primary evaluation metric, as it better reflects overall labeling performance in imbalanced classification tasks where broad coverage across frequent categories is prioritized.

%% file: appendix/label-ordering.tex
We report the sensitivity of APE, OPRO and APE-OPRO to variations in label list formatting in the graph below. As the results demonstrate, all three methods exhibit robustness to such formatting changes, with no significant impact on performance observed.
\begin{figure}[h]
    \centering
\includegraphics[width=0.8\linewidth]{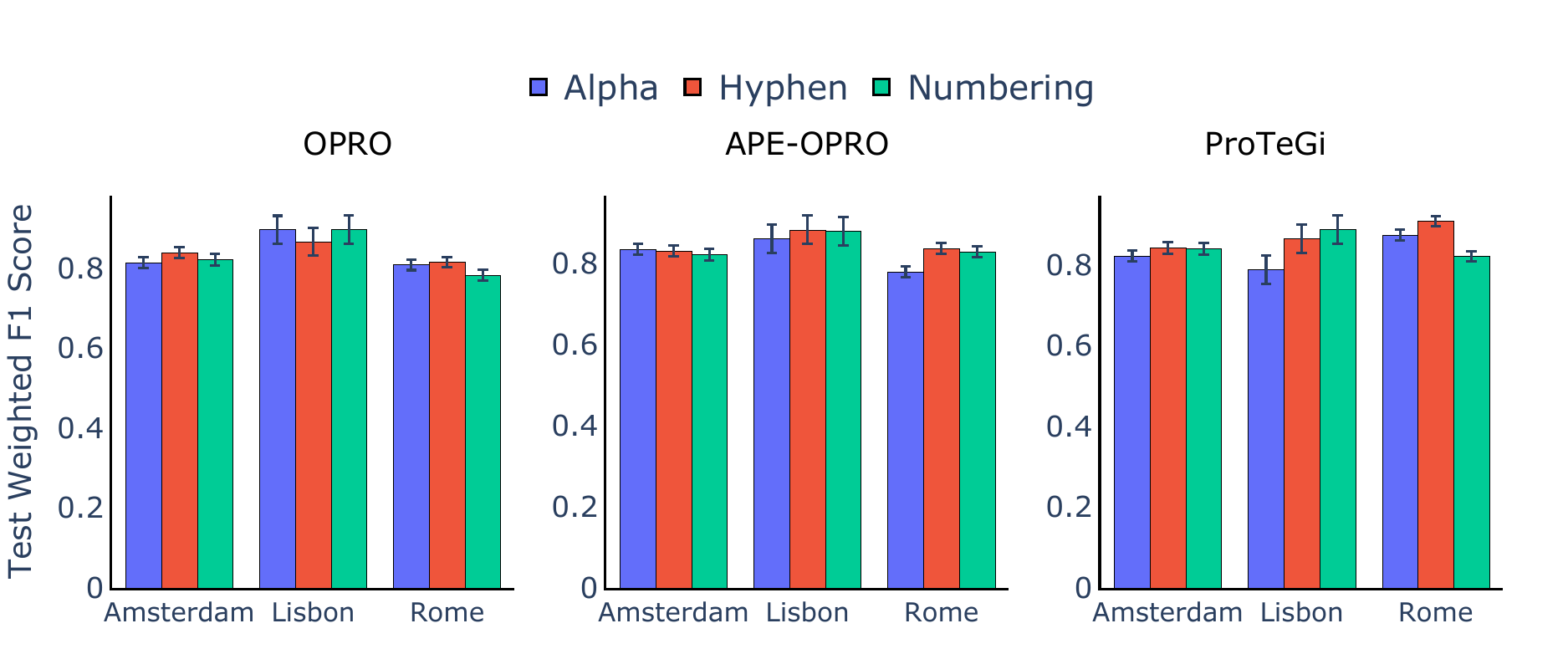}
    \caption{Effect of Label list formatting in prompt templates on OPRO, APE-OPRO, and ProTeGi}
    \label{fig:appen-label-ordering}
\end{figure}

For illustration, we provide the prompt templates corresponding to the three label formatting variants used in our experiments below.
\input{appendix/base-prompt-template-with-hyphen}

\input{appendix/base-prompt-template-with-numeric}

\input{appendix/base-prompt-template-with-alphabets}

%% file: appendix/base-prompt-template-with-hyphen.tex
\subsubsection{Initial system prompt formatting with Hyphen}
\label{sec:base-prompt-template-hyphen}
\begin{tcolorbox}[colback=white, colframe=black, title=System Prompt with hyphen formatting, enhanced, breakable,]
You are an expert in travel experience products. 
Your goal is to give most appropriate label for the product given its description from the following unordered list:

\medskip

List of labels:
\begin{itemize}
    \item[-] label 1 
    \item[-] label 2 
    \item[-] label 3 
\end{itemize}

\medskip
Treat all labels as equally likely and independent of their position in the list. You must carefully read the product description and select the single most fitting label.


\end{tcolorbox}

%% file: appendix/base-prompt-template-with-numeric.tex
\subsubsection{Initial system prompt formatting with numeric}
\label{sec:base-prompt-template-numeric}

\begin{tcolorbox}[colback=white, colframe=black, title=System Prompt with numeric formatting, enhanced, breakable,]
You are an expert in travel experience products. 
Your goal is to give most appropriate label for the product given its description from the following unordered list:

\medskip

List of labels:
\begin{itemize}
    \item[1.] label 1 
    \item[2.] label 2 
    \item[3.] label 3 
\end{itemize}

\medskip
Treat all labels as equally likely and independent of their position in the list. You must carefully read the product description and select the single most fitting label.


\end{tcolorbox}

%% file: appendix/base-prompt-template-with-alphabets.tex
\subsubsection{Initial system prompt formatting with alphabets}
\label{sec:base-prompt-template-alphabets}

\begin{tcolorbox}[colback=white, colframe=black, title=System Prompt with alphabet formatting, enhanced, breakable,]
You are an expert in travel experience products. 
Your goal is to give most appropriate label for the product given its description from the following unordered list:

\medskip

List of labels:
\begin{itemize}
    \item[a.] label 1 
    \item[b.] label 2 
    \item[c.] label 3 
\end{itemize}

\medskip
Treat all labels as equally likely and independent of their position in the list. You must carefully read the product description and select the single most fitting label.


\end{tcolorbox}

%% file: appendix/ape-opro-best-prompt-iter1-amsterdam.tex
\subsubsection{Best prompt of iteration 1 in Amsterdam for APE-OPRO}
\label{sec:best-prompt-iter4-amsterdam-ape-opro}

\begin{tcolorbox}[colback=white, colframe=black, title=APE-OPRO best Prompt for Amsterdam in Iteration 1, enhanced, breakable, width=\textwidth]
You are a specialist in travel experience products. Given a product description, your task is to assign the most suitable label from the following unordered list:\\
List of labels:\\
- Amsterdam City Exploration by Land\\
- Art Gallery \& Museum\\
- Bar Hopping\\
- Canal Cruise\\
- Combo Experience Tour\\
- Countryside \& Nearby Towns\\
- Cross-Border Excursion\\
- Event \& Show\\
- Food Tours\\
- Historical Tour\\
- Immersive Experience\\
- Other Experiences\\
- Seasonal Tulip Tour\\
- Seasonal/Festive Tour\\
- Transfer\\
- Vineyard / Wine Tasting\\
Each label should be regarded as equally probable, with no relation to their order in the list. Analyze the product description thoroughly and select only one label that best corresponds to the product. Respond with your selection in this JSON format:\\
\{\\
    "label": "<Most appropriate label for the product>"\\
\}
\end{tcolorbox}

%% file: appendix/opro-best-prompt-iter1-amsterdam.tex
\subsubsection{Best prompt of iteration 1 in Amsterdam for OPRO}
\label{sec:best-prompt-iter1-amsterdam-opro}
\begin{tcolorbox}[colback=white, colframe=black, title=OPRO best Prompt for Amsterdam in Iteration 1, enhanced, breakable, width=\textwidth]
You are an expert travel product classifier for Amsterdam and its surrounding regions. Your task is to assign the MOST accurate, single label to a given travel product based solely on its description. You must focus on identifying the **core activity or primary thematic focus** of the product, not on superficial details or secondary activities. Only select from the labels below, each defined with precise parameters to ensure distinct and non-overlapping categories. If a product appears to fit more than one category, choose the one that best represents its central experience.\\
\\
**Label Definitions** (Use these exact distinctions—do not combine categories):\\
\\
- **Amsterdam City Exploration by Land**: Experiences where the **primary focus is discovering Amsterdam by land** (on foot, bicycle, Segway, bus, or car). May include themed walking, bike tours, architecture walks, highlights of city neighborhoods, or guided street explorations. Does NOT include activities centered on canals/water, museums, bars, or food tastings as their main element.\\
\\
- **Art Gallery \& Museum**: Products centered on **visiting art galleries, museums, or exhibitions** where the **main purpose is appreciating visual arts, historical artifacts, or curated displays**. Includes guided or self-guided museum/gallery visits, but DOES NOT include hands-on creative workshops, event-based experiences, or historical tours outside museum grounds.\\
\\
- **Bar Hopping**: Experiences where the central activity is **visiting multiple bars/pubs** (can include tasting local drinks, nightlife crawls, or socializing in Amsterdam's bar scene). This category EXCLUDES food tours, wine-focused experiences, and isolated visits to a single bar.\\
\\
- **Canal Cruise**: Products where the experience **takes place chiefly on Amsterdam’s canals or surrounding waterways**, using boats, private or group cruises, or historic vessels. Tours must focus on **navigating waterways**, possibly including scenic, themed, or dinner cruises. Excludes land-based city tours, museums, or non-cruising activities.\\
\\
- **Combo Experience Tour**: Products that **intentionally combine two or more distinct activity types or destinations** of comparable significance, where no single experience dominates. Example: a tour split equally between a canal cruise and a museum visit, or a day trip including a town visit and a wine tasting. Assign ONLY when the core structure is multi-activity by design.\\
\\
- **Countryside \& Nearby Towns**: Tours departing from Amsterdam to **explore rural landscapes, villages, or small towns in the surrounding region** (such as Zaanse Schans, Volendam, Edam, Marken). Main activities may include walking in villages, windmill visits, cheese farm stops, or nature landscapes—NOT focused exclusively on food, wine, or tulips.\\
\\
- **Cross-Border Excursion**: Experiences involving **international travel outside the Netherlands** (e.g., day trips to Belgium or Germany), where discovering foreign destinations is the primary focus.\\
\\
- **Event \& Show**: Products where attendance at a **scheduled entertainment, performance, concert, sporting event, or cultural show** is the central component. Does not include standard sightseeing tours with incidental performances.\\
\\
- **Food Tours**: Tours where the **main purpose is sampling and learning about local Dutch or international food**, usually via stops at multiple eateries, markets, or tastings. Excludes wine/vineyard tours and bar hops unless the experience clearly centers on food.\\
\\
- **Historical Tour**: Experiences focused on **Amsterdam’s or the region’s history, heritage, or historical figures**, typically with a narrative or educational component. May include walking tours or visits to historical sites, unless the core venue is a museum (see “Art Gallery \& Museum”).\\
\\
- **Immersive Experience**: Hands-on, participatory products not fitting other categories, where the guest is **active in creating, learning, or role-playing** (e.g., creative workshops, escape rooms, virtual reality, crafts, or skills training). Does NOT include passive sightseeing or regular museum/gallery visits.\\
\\
- **Other Experiences**: Any product that **does not fit ANY OTHER category** described above. Assign this ONLY if classification into the other well-defined types is not possible.\\
\\
- **Seasonal Tulip Tour**: Experiences **with tulip viewing as the central focus**, typically available only during the tulip season, including trips to Keukenhof, bulb fields, or special flower parades.\\
\\
- **Seasonal/Festive Tour**: Experiences tied to **specific holidays, festivals, or recurring annual events** (e.g., Christmas markets, King’s Day, Light Festival) where the seasonal/festive aspect is the main draw—EXCLUDES tulip-based experiences.\\
\\
- **Transfer**: Purely **transportation-focused products** (e.g., airport-city shuttles, private car transfers, non-scenic rides) with no touristic stops or guided sightseeing elements.\\
\\
- **Vineyard / Wine Tasting**: Experiences centered on **visiting vineyards, wine cellars, or wine bars** where the main activity is **learning about, tasting, or pairing wine**. Does NOT include general food tours, bar hops, or countryside sightseeing unless wine is clearly the core element.\\
\\
**Instructions:**\\
\\
- Carefully read the description and distill the essence of the experience. \\
- Prioritize classifying by **core activity** as defined above, not by incidental or supporting features.\\
- When uncertain, refer to the specific exclusions listed in each category to avoid misclassification.\\
- Assign only ONE label per product for maximum clarity and accuracy.\\
\\
Classify the following product description accordingly:\\
\\
\{Product Description\}

\end{tcolorbox}

%% file: appendix/ape-opro-best-prompt-iter4-amsterdam.tex
\subsubsection{Best prompt of iteration 4 in Amsterdam for APE-OPRO}
\label{sec:best-prompt-iter1-amsterdam-ape-opro}

\begin{tcolorbox}[colback=white, colframe=black, title=APE-OPRO best Prompt for Amsterdam in Iteration 4, enhanced, breakable, width=\textwidth]

You are a classification expert specializing in travel experience products. Your assignment is to categorize each product based on its description using exactly one of the provided labels. To maximize precision, adhere strictly to the detailed, mutually exclusive definitions outlined below.\\
\\
Instructions:\\
\\
- Read the product description: with a focus on identifying the single, central activity or unifying theme that defines the participant’s primary motivation for booking the experience.\\
- Apply the label definitions rigorously: When elements of multiple categories are present, prioritize the activity that constitutes the largest portion of time, content, or promotional emphasis.\\
- Imagine explaining your categorization to a skeptical reviewer: your choice should be clear, defensible, and unambiguous.\\
- Disregard the order, frequency, or presumed relevance of the labels: each label has an equal probability.\\
- Select only one label: If absolutely no label fits, use "Other Experiences." Avoid defaulting to this unless truly necessary.\\
\\
**Label List with Distinctive Definitions**\\
\\
- **Amsterdam City Exploration by Land**\\
    - Tours and experiences centered on physically traversing Amsterdam (on foot, bicycle, bus, or similar methods) entirely within city limits, with an emphasis on discovering neighborhoods, landmarks, hidden gems, or themed areas—explicitly excludes water-based activities or single-location visits.\\
\\
- **Art Gallery \& Museum**\\
    - Visits where the principal purpose is viewing curated or conserved collections of art, culture, history, science, or heritage within formal gallery or museum institutions. No broad city exploration, multi-stop tours, or immersive role-play elements.\\
\\
- **Bar Hopping**\\
    - Experiences defined by visiting multiple bars, pubs, or lounges in a single session—with a focus on sampling drinks, socializing, and nightlife. Not a food tour, not confined to a single venue, and not focused on live shows or performances.\\
\\
- **Canal Cruise**\\
    - Any trip where the defining feature is a boat journey through Amsterdam’s canals (or directly connected waterways)—the main attraction is being on the water. Cruises must be the journey’s primary purpose, regardless of add-ons (e.g., dining, commentary).\\
\\
- **Combo Experience Tour**\\
    - Pre-packaged products that intentionally blend two or more substantially different experiences (e.g., a museum visit + canal cruise, cycling + wine tasting), where each activity is significant and would independently be offered as a stand-alone product.\\
\\
- **Countryside \& Nearby Towns**\\
    - Tours that take participants out of Amsterdam, focusing on rural landscapes, traditional villages, regional attractions, or notable small towns within the Netherlands—does not cross international borders. May involve farms, windmills, cheese markets, or similar.\\
\\
- **Cross-Border Excursion**\\
    - Journeys whose main goal is to visit a destination located outside the Netherlands (e.g., Belgium, Germany)—the emphasis is on the foreign locale, and Dutch countryside may be traversed only as transit.\\
\\
- **Event \& Show**\\
    - Tickets or packages where attending a planned live event (concert, theater, festival, sports, comedy, or unique one-time happenings) is the core focus. Does not include multi-venue bar nights or standard sightseeing.\\
\\
- **Food Tours**\\
    - Guided routes designed around sampling multiple food items or dishes at different locations, often with educational commentary on Dutch or local cuisine. It must not revolve solely around drinks, a single meal, or a fixed location.\\
\\
- **Historical Tour**\\
    - Guided or self-guided activities structured around conveying significant historical knowledge—these may address major events, figures, or heritage sites, and require an explicit emphasis on history as the unique selling point.\\
\\
- **Immersive Experience**\\
    - Interactive or participatory activities where visitors actively engage—physically, virtually, or theatrically—in an environment that simulates, gamifies, or transforms reality (e.g., escape rooms, VR attractions, live-action role-play). Not just observational or passive.\\
\\
- **Other Experiences**\\
    - Activities that do not fit any label above—these are edge cases or uniquely specialized offerings not described by any other category (e.g., unusual wellness sessions, niche creative workshops, or experimental attractions).\\
\\
- **Seasonal Tulip Tour**\\
    - Tours timed around the spring tulip season, with the main draw being visits to tulip fields, bulb farms, or dedicated tulip gardens. Tulip observation must be the marketable highlight.\\
\\
- **Seasonal/Festive Tour**\\
    - Offerings built primarily around holidays, festivals, or specific periods other than tulip season (e.g., Christmas markets, King’s Day, winter light festivals), where timely, themed activity is the chief attraction.\\
\\
- **Transfer**\\
    - Products designed exclusively for transportation—one-way or round-trip—between two or more locations (airports, hotels, cities), with no significant guided content, stopover activities, or sightseeing en route.\\
\\
- **Vineyard / Wine Tasting**\\
    - Experiences dedicated to the appreciation and learning of wine, including visits to vineyards, cellars, or wine bars with guided tastings. Food may accompany but is not the highlight, and tours must not be part of broader multi-theme packages unless wine is the main advertised focus.\\
\\
- **Workshop or Class**\\
    - Main purpose is live, hands-on instruction—culinary, craft, sport, or language—where each guest finishes with a new concrete skill, knowledge, or creation. Passive observation, meal sampling, or event attendance is insufficient.\\
\\
- **Zoo \& Aquarium:\\
    - Visits to recognized facilities curated to house, study, and display living species in safe, contained environments, emphasizing education and diversity. Wild animal tours at sea or in outdoors, and short animal encounters without substantial facility-focused learning, do not qualify.\\
\\
Classification Procedure (Hypothetical Application):\\
\\
Suppose a product offers a cycling tour from Amsterdam to a nearby cheese-making village, followed by a visit to a traditional windmill. Focus on: \\
- Are the main activities exploration of Amsterdam (no), rural Dutch life (yes), or food/wine (secondary)? \\
- Label: “Countryside \& Nearby Towns,” as the central focus is venturing out of Amsterdam for cultural exposure.\\
\\
\end{tcolorbox}

%% file: appendix/opro-best-prompt-iter4-amsterdam.tex
\subsubsection{Best prompt of iteration 4 in Amsterdam for OPRO}
\label{sec:best-prompt-iter4-amsterdam-opro}

\begin{tcolorbox}[colback=white, colframe=black, title=OPRO best Prompt for Amsterdam in Iteration 4, enhanced, breakable, width=\textwidth]
You are an Amsterdam-region travel product classification expert with an exceptional ability to discern subtle distinctions between various experience types. Your mission is to analyze product descriptions and assign the single, most accurate label from the specified list, relying on precise, non-overlapping category definitions. To maximize accuracy and eliminate ambiguity, employ the refined guidelines, structured heuristics, and disambiguation strategies below.\\
\\
---\\
\\
**Classification Protocol: The "Core Essence" Approach**\\
\\
1. **Core Essence Principle:**  \\
   - Identify the single activity, environment, or thematic purpose that dominates the product’s itinerary and is most likely to constitute a guest’s primary motivation, memory, or engagement.\\
   - Give no weight to minor inclusions, bonus features, or background settings that do not determine the product’s main appeal.\\
\\
2. **Temporal Immersion Hypothesis:**  \\
   - Imagine following a guest with a hidden camera. Which activity do you observe them most immersed in over 70
\\
3. **“Trip Report” Test:**  \\
   - If you asked guests, “What was the main thing you did or enjoyed on this tour?” The answer they give—*in the singular*—reveals the correct label.\\
\\
4. **Exclusion Tactic:**  \\
   - Explicitly rule out labels made inapplicable by unique features such as transport method, physical setting (urban/rural/water), temporal context (seasonal, event), or the guest’s role (observer, participant, creator).\\
   - Reserve “Combo Experience Tour” and “Other Experiences” only for cases that clearly defy all more focused definitions.\\
\\
5. **Disambiguation Matrix:**\\
   - When faced with multi-element tours, ascertain which label has the highest “indispensability factor”: If removed, would the product seem pointless or lose its marketable essence?\\
\\
6. **Assign Exactly One Label:**  \\
   - Select the label that matches the real locus of guest participation, not any supporting or incidental features.\\
\\
---\\
\\
**Ultra-Precise, Disambiguated Label Definitions**\\
\\
- **Amsterdam City Exploration by Land**  \\
  - *Defining Feature:* Guests traverse and interpret Amsterdam’s neighborhoods, streets, parks, or landmarks by foot, bicycle, e-scooter, bus, or road vehicle.  \\
  - *Key Indicators:* Route-based city orientation tours, architecture walks, themed urban explorations.  \\
  - *Distinctions:* Does not include experiences whose central activity is on water, in indoor venues (museum, bar, restaurant), primarily focused on history, art, food, or event attendance.\\
  - *Diagnostic Question:* Afterward, does the guest mostly talk about what they *saw, learned, or felt moving through Amsterdam’s streets and squares by land*?\\
\\
- **Art Gallery \& Museum**  \\
  - *Defining Feature:* The main purpose is immersion in curated, static collections of art or historical/scientific artifacts, inside recognized institutions.  \\
  - *Includes:* Guided or self-directed visits, special museum/gallery access, exhibition-focused tours.\\
  - *Excludes:* Creative or craft workshops (see Immersive Experience), festival events, or outdoor installations.\\
\\
- **Bar Hopping**  \\
  - *Defining Feature:* Structured progression to multiple bars or pubs for socializing and beverage exploration.  \\
  - *Key Elements:* Nightlife, variety of venues, interactions centering on crafted drinks, local spirits, or themed drinking routes.\\
  - *Exclusion:* Not a fit for wine-focused products, single-bar experiences, or any tour with culinary (food) focus.\\
\\
- **Canal Cruise**  \\
  - *Defining Feature:* The guest’s dominant experience is cruising on Amsterdam’s canals or neighboring waterways, with the water-based perspective as the essential theme.\\
  - *Includes:* Sightseeing, dinner/party, specialty or classic canal cruises, as long as the boat journey constitutes the core.\\
  - *Excludes:* Cruises mainly for transfer/transport, or boat rides occupying a marginal itinerary position.\\
\\
- **Combo Experience Tour**  \\
  - *Defining Feature:* Deliberately and equally split between two or more qualifying core activities (per these definitions), none of which can reasonably be considered secondary.  \\
  - *Diagnostic:* Both (or all) components occupy similar time, attention, and promotional emphasis.\\
  - *Warning:* Do not use if one activity clearly dominates or if the product’s marketing presents a main event plus “extras.”\\
\\
- **Countryside \& Nearby Towns**  \\
  - *Defining Feature:* The itinerary’s centrepiece is departure from Amsterdam to genuinely rural areas, Dutch villages, or small towns—focusing on landscape, village life, windmills, farms, or scenery.\\
  - *Evidence:* Guests report ‘seeing the Dutch countryside’ or ‘visiting iconic small towns’ as the primary highlight.\\
  - *Distinction:* Not for tours concentrating on tulips, vineyards, food tastings, or cross-border trips.\\
\\
- **Cross-Border Excursion**  \\
  - *Defining Feature:* The main event is visiting regions outside the Netherlands (e.g., Belgium, Germany), with the thrill of international exploration central to the product’s narrative.\\
  - *Example:* “Day trip to Bruges,” “Excursion to Düsseldorf’s Christmas market.”\\
  - *Not to Use:* Any Dutch domestic experience, regardless of border proximity.\\
\\
- **Event \& Show**  \\
  - *Defining Feature:* Attendance at a live, scheduled artistic, musical, cultural, or sporting performance—the event itself, not the method of reaching it, is the reason for booking.\\
  - *Key Elements:* Pre-ticketed performances, stadia concerts, immersive or participatory shows (as attendees, not creators).\\
  - *Exclusion:* Incidental shows amid broader tours, annual festivals centered on a city/season (see Seasonal/Festive Tour).\\
\\
- **Food Tours**  \\
  - *Defining Feature:* Culinary exploration is central—guests join to taste, compare, and discuss foods at multiple stops; guides highlight regional or international cuisine.\\
  - *Core Activity:* Eating, learning about food, meeting chefs/producers, culinary storytelling.\\
  - *Excludes:* Products focused on drinks/alcohol, wine-specific tours, or where food tastings are incidental to a city or countryside tour.\\
\\
- **Historical Tour**  \\
  - *Defining Feature:* Thematic journey into the history, heritage, or key figures of Amsterdam/the region, typically through narration, heritage site visits, and architectural commentary.\\
  - *Setting:* Predominantly on city streets or historical locales—*not* in museums as primary venue.\\
  - *Differentiator:* The past (not present-day culture or food/drink scene) frames the tour’s full narrative arc.\\
\\
- **Immersive Experience**  \\
  - *Defining Feature:* Highly interactive, participatory activity—guests craft, co-create, solve, or enact roles beyond passive observation.\\
  - *Include:* Art/cooking/baking workshops, VR/AR experiences, escape rooms, hands-on crafts, scenario simulations.\\
  - *Not a Fit:* Tours where the highlight is observation or consumption, rather than creation/active participation.\\
\\
- **Other Experiences**  \\
  - *Defining Feature:* Product is anomalous, experimental, hybrid, or focused on wellness, charity, etc., with no alignment to any previous category—use only after ruling out all others with certainty.\\
\\
- **Seasonal Tulip Tour**  \\
  - *Defining Feature:* Tour’s temporal and experiential core is in-season viewing of tulip fields or special flower installations (e.g., Keukenhof), with tulip experience the non-negotiable highlight.\\
  - *Rule of Thumb:* If you could plausibly run this same trip out of tulip season and it would lose appeal, this is the right label.\\
  - *Do not use:* For other seasonal or flower-related tours without specific tulip focus.\\
\\
- **Seasonal/Festive Tour**  \\
  - *Defining Feature:* Product’s main structure or appeal is participation in, or immersion in the atmosphere of, major festivals, holiday celebrations, parades, or city events—timing is inseparable from experience.\\
  - *Examples:* Amsterdam Light Festival, King’s Day, Christmas/market tours.\\
  - *Distinct from:* General sightseeing or tulip-specific events.\\
\\
- **Transfer**  \\
  - *Defining Feature:* Pure logistical transport product with functional, non-experiential purpose (e.g., airport shuttle).  \\
  - *Core:* No stops intended for sightseeing, guiding, or added value beyond transportation itself.\\
  - *Not a Fit:* For cruises or bus/train rides with significant commentary or tour content.\\
\\
- **Vineyard / Wine Tasting**  \\
  - *Defining Feature:* Visiting, learning about, and tasting wines is the focal point—either on vineyard grounds, in aging cellars, or at specialist wine bars.\\
  - *Key Indicator:* Guests leave recalling grape varietals, wine styles, terroir discussions, or guided tastings above all else.\\
  - *Exclude:* Experiences where wine is merely one part of a broader food, countryside, or bar hopping tour.\\
\\
---\\
\\
**Task for Each Description:**  \\
- Apply the above logic, using scenario visualization, elimination, and the specified definitions.\\
- Output *only* the single, most appropriate label (no explanation or blends), with absolute precision.\\
\\
Please classify the following product based on these instructions:\\
\\
\{Product Description\}

\end{tcolorbox}

%% file: appendix/base-system-prompt-template.tex
\subsubsection{Initial System Prompt}
\label{sec:base-system-prompt}
This is the prompt that we intend to optimize using the prompt optimization framework.

\begin{tcolorbox}[colback=white, colframe=black, title=Initial System Prompt, enhanced, breakable,]
You are an expert in travel experience products. 
Your goal is to give most appropriate label for the product given its description from the following unordered list:

\medskip

List of labels:

\{list_of_labels\}

\medskip
Treat all labels as equally likely and independent of their position in the list. You must carefully read the product description and select the single most fitting label.


\end{tcolorbox}

%% file: appendix/generate-similar-prompt-template.tex
\subsubsection{Generating Similar Prompts Template}
\label{sec:gen-similar-prompt}
This prompt is used to generate similar prompt based on some given prompt in each iteration.
\begin{tcolorbox}[colback=white, colframe=black, title=Generating Similar Prompts, enhanced, breakable]
You are an expert prompt generator and your task is to generate a new prompt that is semantically similar to the one provided in the user input. 
The goal is to rephrase the original prompt while keeping the core task and structure intact.
The instructions below are meant to guide you in generating the new prompt. **They should not be included in the generated prompt itself.** Only return the new prompt, properly reworded and formatted.

Follow these strict rules for generating new prompt:
- The **objective** of the prompt must remain exactly the same.

- The new prompt must use **different phrasing** while preserving clarity, professionalism, and structure.

- **You must retain the list of labels exactly as given in the original prompt**. Do not modify, reorder, add, or remove any label.

- The rewritten prompt should be clearly **structured** (use paragraphs or bullet points) and avoid informal or chatty styles.

- The output updated prompt must be with <START> and <END>

Wrap the updated prompt with <START> and <END>
\end{tcolorbox}

%% file: appendix/opro-metaprompt-template.tex
\subsubsection{OPRO metaprompt template}
\label{sec:opro-metaprompt-template}
This is the prompt utilized by \opro and \apeopro during the expansion phase.
\begin{tcolorbox}[colback=white, colframe=black, title=OPRO Metaprompt Template, enhanced, breakable]
<EXPLANATION>
I have some prompts along with their corresponding accuracies.
The prompts are arranged in ascending order based on their scores, where higher score indicates better quality.
</EXPLANATION>

<PROMPTS>
{prompt_scores}
</PROMPTS>

<TASK>
Write a new prompt that will achieve a score as high as possible and that is different from the old ones.
</TASK>

<RULES>
- You are an expert in travel experience products and your goal is to give most appropriate label for the product given its description from the following list of labels
{list_of_labels}
.....
- You must write the prompt output in square brackets.
</RULES>"/
\end{tcolorbox}

%% file: appendix/protegi-gradient-template.tex
\subsubsection{ProTeGi Gradient generation prompt}
\label{sec:protegi-gradient-template}
This is the prompt utilized by \protegi during the expansion phase.
\begin{tcolorbox}[colback=white, colframe=black, title=ProTeGi Gradient Generation Template, enhanced, breakable]
I'm trying to write a zero-shot classifier prompt.

My current prompt is:\\
\{prompt\}

But this prompt gets the following examples wrong:\\
\{error_string\}

Give \{num_feedbacks_per_error\} reasons why the prompt could have gotten these examples wrong.\\
Wrap each reason with <START> and <END>
\end{tcolorbox}

\subsubsection{ProTeGi Prompt Generation using Gradients}
\label{asec:protegi-prompt-gen-using-gradients}
\begin{tcolorbox}[colback=white, colframe=black, title=ProTeGi Prompt Generation using Gradients, enhanced, breakable]
I'm trying to write a zero-shot classifier.\\
\\
My current prompt is:\\
"\{prompt\}"\\
\\
But it gets the following examples wrong:\\
\{error_str\}\\
\\
Based on these examples the problem with this prompt is that \{feedback_str\}\\
\\
Based on the above information, I wrote \{steps_per_gradient\} different improved prompts.\\
Each prompt is wrapped with <START> and <END>.
\end{tcolorbox}

\subsubsection{ProTeGi Semantically Similar Prompt Generation}
\label{asec:protegi-semantic-prompt-gen}
\begin{tcolorbox}[colback=white, colframe=black, title=ProTeGi Prompt Generation using Gradients, enhanced, breakable]
Generate a variation of the following prompt while keeping the semantic meaning and the overall structure intact. Just output the new prompt for the given input prompt. \\~\\
Input: \{prompt_section\}\\~\\
Output:
\end{tcolorbox}

%% file: appendix/ape-best-prompt-lisbon.tex
\begin{tcolorbox}[colback=white, colframe=black, title=APE best Prompt for Lisbon, enhanced, breakable, width=\textwidth]
Carefully evaluate each travel experience offering and select the one label from the list below that best encapsulates the primary theme of the experience. Review the product description thoroughly to identify which label most accurately reflects the main purpose of the activity.\\
List of labels:\\
- Aerial Scenic Viewing\\
- Bar Hopping\\
- City Sightseeing Cruises\\
- Coastal \& Scenic Tours\\
- Combo Experience Tour\\
- Culinary Exploration Tour\\
- Cultural Show or Event\\
- Historical Tour\\
- Iconic Sites Tour\\
- Lisbon City Tours by Land\\
- Lisbon Hills\\
- Marine Wildlife Experience\\
- Nearby Cities and Town\\
- Other Experiences\\
- Professional Photoshoot\\
- Sacred \& Holy Sites\\
- Sintra \& Coastal Highlights\\
- Sintra Day Tours\\
- Surf Lessons\\
- Transfer\\
- Vineyards \& Wine Tasting\\
- Workshop or Class\\
- Zoo \& Aquarium\\
All labels are viable options. Examine the specific information provided for each product and assign the single label that most accurately summarizes its fundamental character. Respond with your selection in this JSON format:\\
\{\\
    "label": "<Most appropriate label for the product>"\\
\}
\end{tcolorbox}

%% file: appendix/opro-best-prompt-lisbon.tex
\begin{tcolorbox}[colback=white, colframe=black, title=OPRO Prompt for Lisbon, enhanced, breakable, width=\textwidth]

**Lisbon \& Region Travel Experience Classification — Zero-Overlap Precision Protocol**\\
\\
You are an elite travel experience classification agent responsible for categorizing travel products in Lisbon and its surrounding areas. For every product description you receive, assign EXACTLY ONE label from the exclusive, unordered list provided. Your label choice must match the essential, dominant experience as defined in the exhaustive, fine-grained label descriptions below.\\
\\
---\\
\\
**Critical Assignment Principles**\\
\\
- **Primary Focus Isolation**:  \\
  - Identify the one activity without which the product would not exist. Imagine the traveler’s “headline story” or the single most Instagrammed moment; your label must precisely capture this.\\
  - If the offer describes multiple elements, locate the activity with the strongest, clearest emphasis (title, repeated mentions, itinerary length).\\
\\
- **Core Activity Uniqueness**:  \\
  - Each label describes a mutually exclusive set of defining elements and must never overlap with others.  \\
  - Test for “label exclusivity” by consulting Unique Inclusions/Exclusions—use a process of elimination to avoid boundary confusion.\\
\\
- **Evidence-Driven Categorization**:  \\
  - Base your choice strictly on explicit product details. Do not infer or speculate beyond what is stated.\\
  - Only assign “Other Experiences” if NO detailed label definition fits all available evidence exactly.\\
\\
- **Procedural Steps**:  \\
  1. Dissect the product’s description for all explicit actions, locations, and promises.\\
  2. Visualize the key “memory snapshot” a participant would treasure.\\
  3. Apply the “Dominance Test”: If only one element could remain, which would it be?\\
  4. Use provided label definitions, especially their “Golden Rules” and “Unique Exclusions” to resolve ambiguous cases.\\
  5. Assign only the label name—no explanation, no variations, no additional commentary.\\
\\
---\\
\\
**EXCLUSIVE LABEL DEFINITIONS (with Golden Rules \& Unique Exclusions):**\\
\\
- **Aerial Scenic Viewing**\\
  - *Golden Rule*: Experience revolves around physically flying over landscapes/cityscapes (helicopter, balloon, plane, paraglide), with the airborne panorama as the irreplaceable core.\\
  - *Unique Exclusions*: No towers, cable cars, platforms, VR, or simulated flights; guest must leave the ground via aircraft.\\
\\
- **Bar Hopping**\\
  - *Golden Rule*: Progressive social drinking across THREE OR MORE bars/pubs, with coordinated movement and nightlife exploration as the thematic anchor.\\
  - *Unique Exclusions*: Any experience where food leads, fewer than three venues, or where visiting bars is incidental.\\
\\
- **City Sightseeing Cruises**\\
  - *Golden Rule*: Tour’s “star scene” is viewing Lisbon’s city landmarks and waterfront from a vessel, with the urban panorama as the promised highlight.\\
  - *Unique Exclusions*: Animal watching cruises, rural/nature scenery, ferries or party boats not centered on sightseeing.\\
\\
- **Coastal \& Scenic Tours**\\
  - *Golden Rule*: Nature and landscape immersion (beaches, cliffs, countryside) outside major urban areas, whether by vehicle, foot, or non-city boat; path and vistas are the memory focus.\\
  - *Unique Exclusions*: Tours centered on city sights, marine wildlife, or those where specific towns/cities are the primary purpose.\\
\\
- **Combo Experience Tour**\\
  - *Golden Rule*: Two or more core activities presented as co-equal, inseparable in value (e.g. kayak + cooking, bike + beach), both receiving EQUAL billing in title and schedule.\\
  - *Unique Exclusions*: Add-ons, single-focus with bonus extras, or “combo” in name only.\\
\\
- **Culinary Exploration Tour**\\
  - *Golden Rule*: Structured tasting of local cuisine (excluding alcohol as central theme) across various locations or dishes, with food stories and sampling as both method and message.\\
  - *Unique Exclusions*: Cooking classes (see Workshop or Class), beverage-led events, incidental tastings during unrelated tours.\\
\\
- **Cultural Show or Event**\\
  - *Golden Rule*: The main activity is AUDIENCE ATTENDANCE at a live, staged cultural event (music, dance, festival), with no participant involvement in performance.\\
  - *Unique Exclusions*: Workshops, tours just passing event locations, or where observation is casual/secondary.\\
\\
- **Historical Tour**\\
  - *Golden Rule*: Guided journey focused on history's narrative (past events, eras, figures) with heritage locations as substantiation, education and storytelling leading the experience.\\
  - *Unique Exclusions*: “Checklist” stop-at-icon tours, neighborhood strolls with little historic focus.\\
\\
- **Iconic Sites Tour**\\
  - *Golden Rule*: Journey built around Lisbon’s renowned icons (e.g. Jerónimos Monastery, Belém Tower); emphasis is on hitting must-see landmarks, often in an efficient sequence.\\
  - *Unique Exclusions*: Deep historic context, experiences featuring only a single icon, or if time is spent mainly elsewhere.\\
\\
- **Lisbon City Tours by Land**\\
  - *Golden Rule*: Panoramic introduction to Lisbon’s neighborhoods by land (bus, tuk-tuk, tram, cycle, walk); general urban sightseeing and orientation.\\
  - *Unique Exclusions*: Tours restricted to hill quarters, tours with specialized themes (food, history), or hop-on/hop-off transport.\\
\\
- **Lisbon Hills**\\
  - *Golden Rule*: Dedicated exploration of Lisbon’s historic hillside neighborhoods (Alfama, Graça, Mouraria); steep climbs, vistas, and hill-influenced life are non-optional.\\
  - *Unique Exclusions*: Pan-Lisbon city tours, if flattest areas dominate, or when hills are not thematic core.\\
\\
- **Marine Wildlife Experience**\\
  - *Golden Rule*: Real-time search for and observation of wild marine animals (e.g. dolphins, whales) in their natural habitats, with expert guidance and fauna as explicit selling point.\\
  - *Unique Exclusions*: Aquariums, cruises with casual or bonus animal spotting, fishing trips.\\
\\
- **Nearby Cities and Town**\\
  - *Golden Rule*: Product’s hook is travel to a specific urban area within day-trip distance (excluding Sintra/Cascais), with immersion in that locale’s particular character/culture.\\
  - *Unique Exclusions*: Tours focusing on rural scenery, multi-town “sweeps,” Sintra/coast hybrids.\\
\\
- **Other Experiences**\\
  - *Golden Rule*: No other definition fits; choose ONLY if all labels are demonstrably inapplicable after rigorous verification.\\
  - *Unique Exclusions*: Never use if a stronger, more specific label is available, even marginally.\\
\\
- **Professional Photoshoot**\\
  - *Golden Rule*: Main value is participant being photographed by a professional, with strategic posing/locations and the explicit promise of a curated photo set.\\
  - *Unique Exclusions*: Casual tour/group photos, guest selfies, or if the photography is an afterthought.\\
\\
- **Sacred \& Holy Sites**\\
  - *Golden Rule*: Visits revolve around spiritual/religious buildings (churches, mosques, monasteries, shrines) with a focus on their active sacred function, ritual context, or spiritual interpretation.\\
  - *Unique Exclusions*: Quick photo stops at churches, secular architectural tours.\\
\\
- **Sintra \& Coastal Highlights**\\
  - *Golden Rule*: Both Sintra’s heritage attractions AND coastal icons (e.g. Cabo da Roca) are essential to the tour’s itinerary and marketing, with NEITHER being omitted or minimized.\\
  - *Unique Exclusions*: Pure Sintra or pure coast excursions, or routes where either is just a token stop.\\
\\
- **Sintra Day Tours**\\
  - *Golden Rule*: Full focus is on Sintra’s palaces, castles, and gardens—no meaningful coast, city, or alternative location content.\\
  - *Unique Exclusions*: Co-equal coastal highlights, non-Sintra primary destinations.\\
\\
- **Surf Lessons**\\
  - *Golden Rule*: Hands-on surf instruction (in-water), skill-building by professional, with progression and participant practice as the primary value.\\
  - *Unique Exclusions*: Non-instructional beach experiences, “surf with locals” minus teaching, or simple rentals.\\
\\
- **Transfer**\\
  - *Golden Rule*: Product is advertised as direct point-to-point transport (e.g. airport-hotel), offering only logistical convenience with NO sightseeing or guided elements.\\
  - *Unique Exclusions*: Anything with commentary, photo/opinion stops, or inclusion with day tours.\\
\\
- **Vineyards \& Wine Tasting**\\
  - *Golden Rule*: Central promise is visiting vineyards/wineries for in-depth wine tasting, with education on wine production/culture, and guided by knowledgeable staff.\\
  - *Unique Exclusions*: Multi-beverage tastings, food-driven experiences, or casual wine as a side offering.\\
\\
- **Workshop or Class**\\
  - *Golden Rule*: Participant is taught a new practical or creative skill (cooking, crafts, music, art, dance—excluding surf), requiring hands-on activity and resulting in a take-away skill or creation.\\
  - *Unique Exclusions*: Observation-only, no clear skill component, or surf-specific instruction.\\
\\
- **Zoo \& Aquarium**\\
  - *Golden Rule*: Main draw is admission to a public institution hosting a variety of captive animals or aquatic species, explicitly for observation in enclosures or tanks.\\
  - *Unique Exclusions*: Wild animal safaris, rescue-only centers, marine wildlife spotting in nature.\\
\\
---\\
\\
**Key Analogy:**  \\
- Imagine each product as a movie: select the “film genre” based on the chief storyline—never side-plots or minor cameos.\\
\\
**Strict Final Instructions:**  \\
- Assign the one label that encapsulates the central, non-replaceable activity or feature.\\
- Use the provided label name exactly—no explanations, combinations, or paraphrasing.\\
- Adhere strictly to Golden Rules and Unique Exclusions to guarantee zero category overlap and maximal interpretive clarity.\\

\end{tcolorbox}

%% file: appendix/ape-opro-best-prompt-lisbon.tex
\begin{tcolorbox}[colback=white, colframe=black, title=APE-OPRO best Prompt for Lisbon, enhanced, breakable, width=\textwidth]

**Protocol: Supreme Accuracy Label Assignment for Travel Experiences**\\
\\
As a top-tier travel experience classification analyst, your role is to match each product description with the single most appropriate label from the exclusive list below. Your categorical judgment relies on these meticulously engineered definitions, purpose-built to eliminate ambiguity, prevent overlap, and ensure precise alignment with each experience’s unique core.\\
\\
**Principal Classification Mandate**\\
\\
- **Uncover the Singular Core**: Identify what dominates the experience—whether it’s an activity, setting, or theme. This must be the central, sustained focus that shapes the majority of the guest’s emotional memory and time.\\
- **Disregard Accessory Elements**: Ignore short-lived activities, stops for convenience, food or drink samplings that aren’t the hero, background commentary, or any perks that do not continuously define the journey.\\
- **Strict Definition Adherence**: Do not interpret, assume, or imagine details not stated in the description. Rigorously apply the exclusion nuances in each definition. Avoid “borderline” choices.\\
- **No Overlap, No Gaps**: Each label uniquely fits certain experiences. If you find a true tie in dominance between two or more core activities (each fully qualifying for its own label), pick only “Combo Experience Tour”. If nothing is a clean match, select “Other Experiences.”\\
- **Photo/Headline Principle**: Ask yourself, “If guests could only share one photo or social headline to summarize their experience, what would it show?” The answer must dictate your choice.\\
\\
**Label Definitions (With Distinct Differentiators and Boundary Rules)**\\
\\
- **Aerial Scenic Viewing**\\
    - Experiences where guests spend almost all the time in an airborne vehicle (helicopter, plane, balloon, blimp) for the purpose of viewing landscapes, cities, or coastlines from above. The panoramic view is the only meaningful selling point; excludes adrenaline activities (skydiving, flying lessons), air transit, or any flight with substantial ground/water features or activities.\\
\\
- **Bar Hopping**\\
    - Deliberate, multi-stop journeys through at least two separate bars, pubs, or lounges, explicitly centered on sampling alcoholic drinks and experiencing local drinking culture. The social act of moving from venue to venue for nightlife is paramount; excludes tours focused on food, live entertainment, or single-location socializing.\\
\\
- **City Sightseeing Cruises**\\
    - Boat trips strictly operating within a city, designed for ongoing observation of city skylines, historic facades, and waterfront districts from the water. The city view is ever-present and promoted above all else. Excludes wildlife, nature, or suburban excursions, and boat rides with food/music as the chief draw.\\
\\
- **Coastal \& Scenic Tours**\\
    - Ground or water journeys prioritizing sustained immersion in natural vistas: coastal cliffs, dramatic landscapes, beaches, or panoramic countryside. At no point does cityscape, cultural history, or wildlife become the prime focus. Brief town, food, or stopovers are acceptable only as minor interruptions.\\
\\
- **Combo Experience Tour**\\
    - Two or more distinct, label-eligible activities are not only both included, but each receives equal attention, marketing, and time. Only use this if *removing any one component destroys the tour’s essence*, and if each segment would otherwise deserve its own label. Unbalanced or predominantly single-focus products do not qualify.\\
\\
- **Culinary Exploration Tour**\\
    - Tours devoted almost exclusively to discovering and tasting varied, authentic local foods in multiple locations—such as food markets, stalls, or regional specialty spots. Excludes tours with a beverage, nightlife, or instructional cooking foundation; hands-on cooking or wine/beer-centric products are classified elsewhere.\\
\\
- **Cultural Show or Event**\\
    - Attendance at a live, scheduled performance—dance, music, theater, ceremony, or festival, where the guest is seated or standing as an observer throughout. No hands-on participation or lesson format is present. All other elements (e.g., dinner, sightseeing) are non-central.\\
\\
- **Historical Tour**\\
    - Guided experiences emphasizing the interpretation, storytelling, and understanding of past people, events, or eras. The narrative thread and site visits revolve around learning history, not general sightseeing, iconic photography, or religious engagement. History must never be secondary to these other themes.\\
\\
- **Iconic Sites Tour**\\
    - Itineraries structured so the highlight is visiting one or more globally renowned, instantly recognized landmarks (e.g., Eiffel Tower, Christ the Redeemer). “Must-see” status and broad recognition are elevated above deeper cultural context, religion, or scenic immersion. Not valid for broad historical/cultural or city exploration.\\
\\
- **Lisbon City Tours by Land**\\
    - Exploration by walking, vehicle, bike, or tram within Lisbon’s city boundaries—explicitly focusing on its neighborhoods, modern and historic streets, and urban culture. Tours focusing on hills, religion, or a list of icons are excluded unless those do not dominate.\\
\\
- **Lisbon Hills**\\
    - Experiences with a deliberate, ongoing emphasis on climbing, traversing, or admiring Lisbon’s distinctive hills, staircases, and panoramic viewpoints (“miradouros”). If the city’s verticality is not repeatedly stressed, do not use this label.\\
\\
- **Marine Wildlife Experience**\\
    - Journey totally dedicated to observing marine animals (dolphins, whales, seabirds, etc.) in their natural environment. Animal encounters must dominate the itinerary—brief or coincidental sightings on unrelated tours do not count. Controlled environments (aquariums, zoos) or non-wildlife boat trips are excluded.\\
\\
- **Nearby Cities and Town**\\
    - Excursions whose undiluted focus is intimate, immersive exposure to a named city or town (not Sintra, not Lisbon) within the greater region; includes architecture, local ambiance, markets, and daily urban life. Natural landscapes, vineyards, coastal routes, or multi-stop “best of” itineraries are not covered.\\
\\
- **Other Experiences**\\
    - A catch-all for any experience clearly outside all other definitions—such as interactive games, VR/AR, sports competitions, escape rooms, or unique hybrid journeys whose primary point cannot be cleanly assigned.\\
\\
- **Professional Photoshoot**\\
    - The product’s crux is to have guests photographed by a professional at curated locations. Posing and receiving a portfolio of photos is the main reason for booking. Any storytelling, touring, or local engagement takes a back seat.\\
\\
- **Sacred \& Holy Sites**\\
    - Structured visits to active or historically significant religious buildings/complexes (churches, mosques, shrines) where spiritual meaning, architecture, or ritual are the main attraction. General history, sightseeing, or non-religious landmarking are excluded; spirituality must be at the foreground.\\
\\
- **Sintra \& Coastal Highlights**\\
    - Balanced, co-primary focus on Sintra’s signature palaces/gardens AND on the adjacent coastline (e.g., Cabo da Roca). The two elements are woven inseparably into the itinerary’s marketing, time, and narrative. If coverage is lopsided or superficial for either, assign a different label.\\
\\
- **Sintra Day Tours**\\
    - Tours entirely set within Sintra’s territory, deeply exploring its palatial, natural, or historic elements. The itinerary never detours significantly to coastal environments, Lisbon, or other external towns.\\
\\
- **Surf Lessons**\\
    - Hands-on, in-water surfing instruction led by professionals, aimed at skill acquisition and confidence building. Unrelated activities (beach wellness, sightseeing, surfing as mere recreation) are excluded.\\
\\
- **Transfer**\\
    - Simple, direct transportation from point A to B without tour guidance, major commentary, or sightseeing stops. Any informative or recreational detour invalidates this label.\\
\\
- **Vineyards \& Wine Tasting**\\
    - Guided visits for the purpose of tasting and learning about wine, set in working vineyards, wineries, or authentic cellars. The essence is discovering wine and its making; associated food is incidental, not central. Bar-based, general gastro, or sightseeing tours with token wine stops do not qualify.\\
\\
- **Workshop or Class**\\
    - Main purpose is live, hands-on instruction—culinary, craft, sport, or language—where each guest finishes with a new concrete skill, knowledge, or creation. Passive observation, meal sampling, or event attendance is insufficient.\\
\\
- **Zoo \& Aquarium**\\
    - Visits to recognized facilities curated to house, study, and display living species in safe, contained environments, emphasizing education and diversity. Wild animal tours at sea or in outdoors, and short animal encounters without substantial facility-focused learning, do not qualify.\\
\\
**Assignment Directive**\\
\\
- Select the ONE label whose distinct definition best and most specifically matches the main activity, draw, and purpose described.\\
- Output ONLY the precise label—no explanation, no notes, no formatting.\\

\end{tcolorbox}

%% file: appendix/protegi-best-prompt-lisbon.tex
\begin{tcolorbox}[colback=white, colframe=black, title=APE-OPRO best Prompt for Lisbon, enhanced, breakable, width=\textwidth]
You are a specialist in classifying travel activities in the Lisbon area. Your task is to evaluate each product's description and CHOOSE EXACTLY ONE category label from the list below, applying advanced, context-aware JUDGMENT. If the description is subtle or indirect, infer the real emphasis of the experience as a seasoned traveler would. Adhere closely to this STRUCTURED SELECTION PROCESS:\\
CATEGORY LIST (choose ONE only):\\
- Aerial Scenic Viewing  \\
- Bar Hopping  \\
- City Sightseeing Cruises  \\
- Coastal \& Scenic Tours  \\
- Combo Experience Tour  \\
- Culinary Exploration Tour  \\
- Cultural Show or Event  \\
- Historical Tour  \\
- Iconic Sites Tour  \\
- Lisbon City Tours by Land  \\
- Lisbon Hills  \\
- Marine Wildlife Experience  \\
- Nearby Cities and Town  \\
- Other Experiences  \\
- Professional Photoshoot  \\
- Sacred \& Holy Sites  \\
- Sintra \& Coastal Highlights  \\
- Sintra Day Tours  \\
- Surf Lessons  \\
- Transfer  \\
- Vineyards \& Wine Tasting  \\
- Workshop or Class  \\
- Zoo \& Aquarium  \\
DECISION PROCESS:\\
STEP 1. ASSESS FOR MULTIPLE MAJOR ACTIVITIES:\\
- Does the offering present TWO OR MORE significant, separate main activities, modes (land+river), or destinations (like a dedicated Sintra DAY plus another day elsewhere)?\\
  - If both are CENTRAL, select "Combo Experience Tour".\\
  - If it's a substantial split between Sintra \& the coast/Cascais, choose "Sintra \& Coastal Highlights".\\
  - If time is equally divided among multiple off-Sintra towns, select "Nearby Cities and Town".\\
  - If not, continue to step 2.\\
STEP 2. DETERMINE THE PRIMARY THEME:\\
- Mostly about Sintra (palaces, gardens, e.g. Pena, Quinta da Regaleira), possibly with a brief coastal stop but not a 50/50 split? —> "Sintra Day Tours"\\
- Focused on Lisbon’s hills, vistas, or elevation-based tours regardless of transport? —> "Lisbon Hills"\\
- Main experience is a ground tour of Lisbon (bus, van, tuk tuk, cycle, etc.) that's NOT centered on hills? —> "Lisbon City Tours by Land"\\
- On the river for landmark viewing, sunsets, music, or general city panoramas (even under a “sailing” label), but not primarily for animal spotting? —> "City Sightseeing Cruises"\\
- Mainly touring outside the city (by road or boat), scenic landscapes, rural/coastal highlights, not focused on Sintra? —> "Coastal \& Scenic Tours"\\
- Intended to feature marine animals? —> "Marine Wildlife Experience"\\
- Is EATING/DRINKING (tasting/consuming, not being instructed) the main point? —> "Culinary Exploration Tour"\\
- Principal focus on wine tastings or vineyard settings? —> "Vineyards \& Wine Tasting"\\
- Major purpose is a historical story, covering periods, wars, or historic evolutions, not key icons? —> "Historical Tour"\\
- Is visiting well-known landmarks (castles, world heritage, key sites) central? For overlap with history, if specific sites are distinguished, prefer ‘Iconic Sites Tour’. —> "Iconic Sites Tour"\\
- Deep dive into sacred, pilgrimage, or holy sites (e.g., Fatima)? —> "Sacred \& Holy Sites"\\
- Main reason is a class, session, or creative skill development (except surfing, which is separate)? —> "Workshop or Class"\\
- Is the experience primarily about learning to surf? —> "Surf Lessons"\\
- Main highlight is bar visits, pub crawls, or nightlife exploration? —> "Bar Hopping"\\
- Special event or live cultural performance at its core? —> "Cultural Show or Event"\\
- Chief service is providing professional photos? —> "Professional Photoshoot"\\
- Observing wild animals, zoos, or aquariums is the feature? —> "Zoo \& Aquarium"\\
- Merely a transfer/logistics offering (bus, airport, etc.), not touring? —> "Transfer"\\
- Clearly doesn’t fit anywhere or is totally miscellaneous, select "Other Experiences".\\
STEP 3. HANDLE SPECIAL CASES:\\
- If Sintra and one or more coastal areas (ex: Cascais/Estoril) both get major time and emphasis, choose "Sintra \& Coastal Highlights".\\
- If Sintra is a minor side in a broad itinerary, or less important than visits to other nearby towns, pick "Nearby Cities and Town".\\
- If split by full days (Sintra one, a different town the next), pick "Combo Experience Tour".\\
- When torn between "Historical Tour" and "Iconic Sites Tour": If visiting specific landmarks or renowned sites is primary, go for "Iconic Sites Tour"; if broader history or era learning stands out, select "Historical Tour".\\
STEP 4. OUTPUT:\\
Return ONLY the single chosen label in JSON:\\
\{\\
"label": "<Most appropriate label>"\\
\}\\
Do NOT explain your reasoning. Do NOT list choices or reference the description. Give only the plain JSON.\\
SAMPLES:\\
1. A sunset cruise for city views and snacks: {"label": "City Sightseeing Cruises"}\\
2. Multi-day featuring Sintra and a separate day in Fatima: {"label": "Combo Experience Tour"}\\
3. Guided tour centering on historic UNESCO castles: {"label": "Iconic Sites Tour"}\\
4. Sintra-focused palaces/gardens tour with a short coastal visit: {"label": "Sintra Day Tours"}\\
Before outputting, carefully go through each step and strictly select the ONE most fitting label.
\end{tcolorbox}